%% file: main.tex
\documentclass[10pt,twocolumn,letterpaper]{article}

\usepackage{iccv}              % To produce the CAMERA-READY version
\usepackage[accsupp]{axessibility} % Improves PDF readability for those with disabilities.
\usepackage{multirow}
\usepackage{multicol}
\newcommand{\tabincell}[2]{\begin{tabular}{@{}#1@{}}#2\end{tabular}}
\usepackage{graphicx}
\usepackage{float}
\usepackage{afterpage}
\usepackage{subfiles}


\definecolor{iccvblue}{rgb}{0.21,0.49,0.74}
\usepackage[pagebackref,breaklinks,colorlinks,allcolors=iccvblue]{hyperref}

\def\paperID{8217} % *** Enter the Paper ID here
\def\confName{ICCV}
\def\confYear{2025}

\title{Diving into the Fusion of Monocular Priors for Generalized Stereo Matching}
\author{
Chengtang Yao\textsuperscript{1,2}, Lidong Yu\textsuperscript{3}, Zhidan Liu\textsuperscript{1,2}, Jiaxi Zeng\textsuperscript{1,2}, Yuwei Wu\textsuperscript{1,2}\footnotemark[1], Yunde Jia\textsuperscript{2,1}\footnotemark[1]\\
\textsuperscript{1}Beijing Key Laboratory of Intelligent Information Technology,\\
School of Computer Science \& Technology, Beijing Institute of Technology, China\\
\textsuperscript{2}Guangdong Laboratory of Machine Perception and Intelligent Computing,\\
Shenzhen MSU-BIT University, China\\
\textsuperscript{3}NVIDIA\\
{\tt\small \{zdliu, wuyuwei, jiayunde\}@bit.edu.cn} \\
{\tt\small \{yao.c.t.adam, yvlidong, jiaxizeng.jx\}@gmail.com}
}

\begin{document}
% \maketitle
% \twocolumn[
% \maketitle

% % 插入跨双栏大图
% \afterpage{%
%     \begin{figure*}[htbp]
%         \centering
%         \includegraphics[width=\textwidth]{Figure/fig1-all.pdf}
%         \caption{The visualization of different ill-posed areas in the Booster dataset.}
%         \label{fig:fig1}
%     \end{figure*}
%     \clearpage
% }
% ]

% \begin{figure*}[htbp]
% \centering
%     \includegraphics[width=.95\textwidth]{Figure/fig1-all.pdf}
%     % \vspace{-0.55cm}
%     \caption{The visualization of different ill-posed areas on Booster dataset.}
%     \vspace{-0.3cm}
%     \label{Fig: vis booster}
% \end{figure*}

% \maketitle

\twocolumn[
\maketitle
{
\vspace{-12mm}
\begin{center}
  \href{https://huggingface.co/spaces/AdamYao/Diving-into-the-Fusion-of-Monocular-Priors-for-Generalized-Stereo-Matching}{%
    \includegraphics[height=12pt]{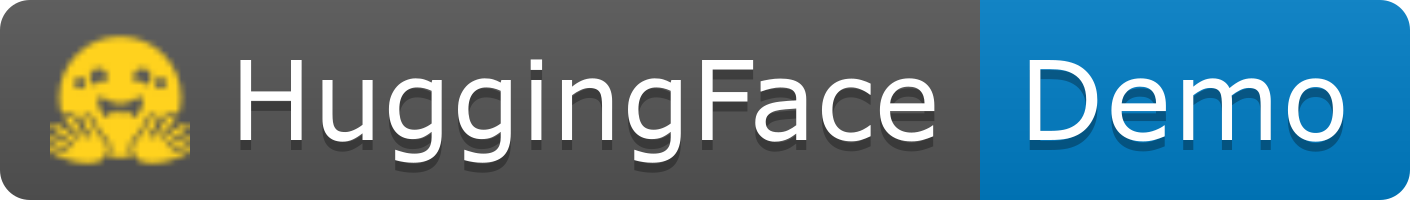}
  }%
  \hspace{0.8em}%
  \href{https://github.com/YaoChengTang/Diving-into-the-Fusion-of-Monocular-Priors-for-Generalized-Stereo-Matching}{%
    \includegraphics[height=12pt]{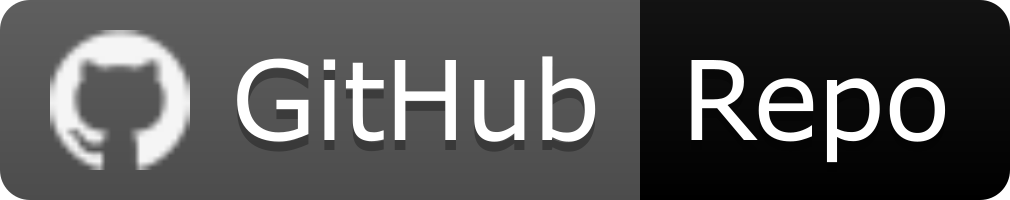}
  }%
  \hspace{0.8em}%
  \href{https://drive.google.com/drive/folders/1PaQPOzzDajlnFfNm2fBKZsfJKb2XPejd?usp=sharing}{%
    \includegraphics[height=12pt]{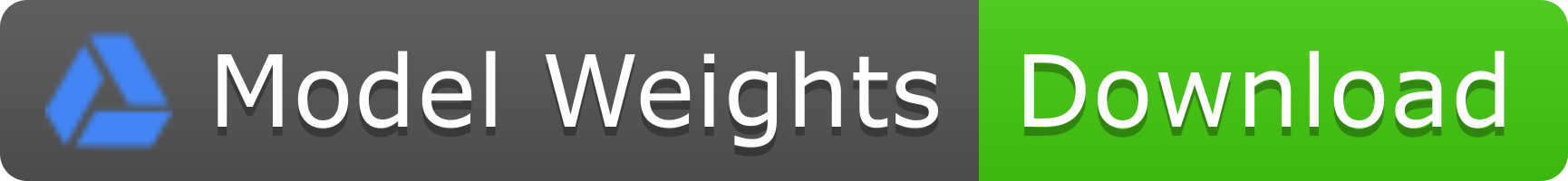}
  }
\end{center}
% \vspace{-2mm}
}]

\renewcommand{\thefootnote}{\fnsymbol{footnote}}
\footnotetext[1]{Corresponding author.}

% \addtocontents{toc}{\protect\setcounter{tocdepth}{-10}}

\input{sec/0_abstract}
\vspace{-3mm}
\input{sec/1_intro}
\input{sec/2_relatedwork}
\input{sec/3_method}
\input{sec/4_experiments}
\input{sec/5_conclusion}

{
    \small
    \bibliographystyle{ieeenat_fullname}
    % \bibliography{main}
    \bibliography{depth}
}

% \addtocontents{toc}{\protect\setcounter{tocdepth}{2}}%
\input{sec/6_supp}

\end{document}

%% file: sec/0_abstract.tex
\begin{abstract}
% \vspace{-2mm}
The matching formulation makes it naturally hard for the stereo matching to handle ill-posed regions like occlusions and non-Lambertian surfaces. Fusing monocular priors has been proven helpful for ill-posed matching, but the biased monocular prior learned from small stereo datasets constrains the generalization. Recently, stereo matching has progressed by leveraging the unbiased monocular prior from the vision foundation model (VFM) to improve the generalization in ill-posed regions. We dive into the fusion process and observe three main problems limiting the fusion of the VFM monocular prior. The first problem is the misalignment between affine-invariant relative monocular depth and absolute depth of disparity. Besides, when we use the monocular feature in an iterative update structure, the over-confidence in the disparity update leads to local optima results. A direct fusion of a monocular depth map could alleviate the local optima problem, but noisy disparity results computed at the first several iterations will misguide the fusion. In this paper, we propose a binary local ordering map to guide the fusion, which converts the depth map into a binary relative format, unifying the relative and absolute depth representation. The computed local ordering map is also used to re-weight the initial disparity update, resolving the local optima and noisy problem. In addition, we formulate the final direct fusion of monocular depth to the disparity as a registration problem, where a pixel-wise linear regression module can globally and adaptively align them. Our method fully exploits the monocular prior to support stereo matching results effectively and efficiently. We significantly improve the performance from the experiments when generalizing from SceneFlow to Middlebury and Booster datasets while barely reducing the efficiency.

\end{abstract}

%% file: sec/1_intro.tex
% \begin{figure}[htbp]
%     \centering
%     \includegraphics[width=.4\textwidth]{Figure/fig1_2.pdf}
%     % \vspace{-0.55cm}
%     \caption{The domain generalized results of stereo matching methods training on SceneFlow dataset and testing on Booster dataset. Our method shows great in-the-wild generalization ability on the ill-posed regions, while state-of-the-art methods fall into local optima in the blue box areas.}
%     \vspace{-0.4cm}
%     \label{fig:fig1}
% \end{figure}

\section{Introduction}
\label{sec:intro}

Stereo matching provides dense depth for various downstream applications, such as autonomous driving, robotics, AR/MR, etc. These applications require stereo matching to generalize across different scenes from wild worlds. However, the generalization of stereo matching becomes poor in ill-posed regions due to occlusion, texture-less, and non-Lambertian surfaces (e.g., reflective or transparent surfaces). Fusion of monocular priors is proven to help correct the ill-posed binocular matching results \cite{jing2023uncertainty,lou2023elfnet,xu2023iterative,rao2023masked,liu2022graftnet,chuah2022itsa,zhang2022revisiting,li2024local}. But the monocular prior trained on the limited data distribution of stereo datasets is susceptible to domain bias and can only capture significantly biased monocular features for certain scenes \citep{guo2017calibration,ovadia2019can}.

\begin{figure*}[htbp]
\vspace{-0.5cm}
\begin{center}
{
    \includegraphics[width=1.\textwidth]{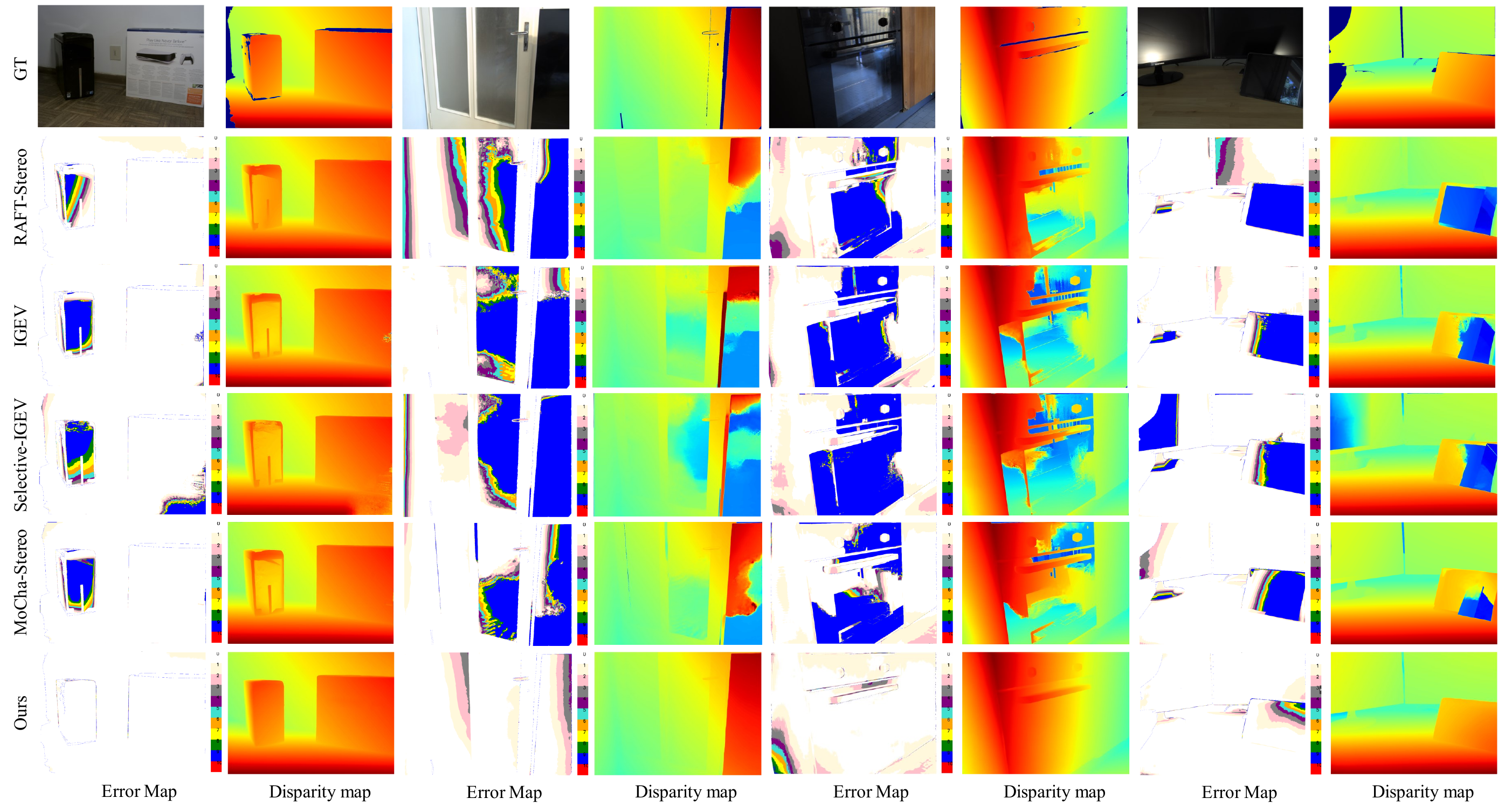}
}
\vspace{-0.5cm}
\captionof{figure}{The visualization of different ill-posed regions in the Booster dataset. Our method achieves an overwhelming advantage in all kinds of regions.}
\label{fig: fig1}
\end{center}
\vspace{-6mm}
\end{figure*}

Taking advantage of large-scale scenes and the easily collected ground truth of monocular depth, the vision foundation model can provide an unbiased monocular prior \cite{ke2024repurposing,yang2024depth,yang2024depthv2}. Recently, some methods have made great progress in fusing the unbiased monocular prior into the stereo matching to improve the generalization in ill-posed regions \cite{cheng2025monster, bartolomei2024stereo, wen2025foundationstereo}. In this paper, we dive into the fusion mechanism and find three main problems limiting a full exploration of the unbiased monocular prior. The first problem lies in the natural gap between the affine-invariant relative depth from monocular depth and absolute depth from disparity. Although we can forcibly align the two kinds of depth with a complex mutual refinement, these alignments could involve heavy computation and greatly harm the efficiency \cite{cheng2025monster, wen2025foundationstereo}. The other problem exists in the fusion with monocular feature maps in an iterative refinement structure \cite{xu2023iterative,wang2024selective,chen2024mocha}. The implicit feature fusion makes the fusion more biased to the binocular information due to the iterative update training scheme, where the over-confidence of the disparity update causes local optima, as shown in Figure \ref{fig: fig1}.  
An additional fusion of monocular depth could alleviate local optima, but direct fusion of the depth is easily affected by noisy depth results. Even with unbiased, smooth monocular depths from the VFM, noisy disparity in the first several iterations slows down good fusion.
% An additional fusion of monocular depth could alleviate the local optima, but the direct fusion of the depth map is easily affected by the noisy depth results. Even with unbiased, smooth monocular depths from the VFM, the noisy disparity at the first several iterations slows down a good fusion.

In this paper, we present a new depth representation called the local binary ordering map that indicates whether two pixels are farther or closer. It converts the depth into a binary relative depth representation, unifying the monocular depth and binocular disparity. The local binary order map also guides fusion in an explicit manner, which restricts the influence of the large noise from outliers. Furthermore, we formulate the binocular disparity map as a noisy version of monocular depth registered by specific pixel-wise scale and shift. Therefore, the alignment between monocular depth and binocular disparity can be deemed a noisy linear regression problem about the registration parameters. The registration formulation globally and adaptively aligns the two kinds of depth in an efficient manner.

Our network can be divided into three modules. The monocular encoder extracts unbiased monocular priors, including monocular depth and context features, using a large pre-trained monocular network like \cite{yang2024depth,yang2024depthv2,yin2023metric3d}. Then, the fusion can be realized by an iterative local fusion module and a global fusion module to fully exploit the usage of the monocular priors with matching information. The iterative local fusion module uses a two-stream architecture to update the disparity iteratively. The first stream computes two binary ordering maps from monocular depth and binocular disparity through a series of LBP-like convolution blocks. Then, we compute the differences between the two binary ordering maps to form a local guidance for fusion. At the same time, the second stream predicts an initial disparity update result through a multi-level GRU using cost volume and monocular context features. The local guidance is used to re-weight the initial disparity update result, resolving the local optima. After local fusion, the global fusion module realizes the optimization of the disparity map by registering to monocular depth. We first compute two parameters to register the relative depth to the absolute depth globally. It solves the noisy linear regression problem between optimized disparity and monocular depth through a series of convolutions. Then, we compute a confidence map using the cost volume, the hidden state of GRU, and the local guidance from the last iteration. The confidence map guides the fusion of the optimized binocular disparity and the registered monocular depth as the final prediction.

% We compare our model with SOTA methods using the standard setting, training on the SceneFlow dataset, and testing on five real-world datasets with various ill-posed areas, including KITTI 2012, KITTI 2015, Middlebury, ETH3D, and Booster. The results demonstrate that our method significantly enhances the performance of SOTA approaches, as shown in Figure \ref{fig: fig1}. Experiments show that our method achieves a 10-point improvement in the bad2 metric for transparent regions on the Booster dataset and reduces errors by more than 50\% on Middlebury and ETH3D, where we do not use additional stereo data or specific data augmentation. Meanwhile, even involving a VTF model, our method barely raises the time cost, benefiting from the elegant explicit designs.

We compare our model with SOTA methods under the standard setting: training on SceneFlow and testing on five real-world datasets (KITTI 2012\&2015, Middlebury, ETH3D, and Booster) with various ill-posed regions. Results demonstrate our method significantly boosts SOTA performance, as shown in Figure \ref{fig: fig1}. 
Our method achieves a 10-point improvement in the bad2 metric for transparent regions on Booster and reduces errors by over 50\% on Middlebury and ETH3D, without using extra stereo data or specific augmentation. Despite involving a VTF model, it barely increases the time cost due to the elegant designs.
% Our method achieves a 10-point improvement in the bad2 metric for transparent regions on Booster and reduces errors by over 50\% on Middlebury and ETH3D, without using extra stereo data or specific augmentation. Our method barely increases the time cost even involving a VTF model due to the elegant explicit designs.

%% file: sec/2_relatedwork.tex
\section{Related Work}
\label{sec:relatedwork}

\subsection{Generalized Stereo Matching}
% Generalized stereo matching aims to produce a reliable dense disparity map when the target domain (e.g., real-world data) differs from the source domain (e.g., synthetic data). Some methods focus on the learning of domain invariant features \citep{cai2020matching,liu2022graftnet,chuah2022itsa,zhang2022revisiting,rao2023masked,kim2022pointfix,tonioni2019learning}. MS-PSMNet \citep{cai2020matching} replaces the learning-based features with hand-crafted features to force the stereo network focus on the matching space. MS-PSMNet has achieved great improvement, but its hand-crafted features limit the performance of the stereo network. Thus, many methods turn to improve the training process by transforming learning \citep{liu2022graftnet}, meta-learning \citep{tonioni2019learning,kim2022pointfix}, contrastive learning \citep{zhang2022revisiting,rao2023masked}, and fisher information \citep{chuah2022itsa}.
Generalized stereo matching aims to produce a reliable dense disparity map when the target domain (e.g., real-world data) differs from the source domain (e.g., synthetic data). Some methods focus on learning domain-invariant features \citep{cai2020matching,liu2022graftnet,chuah2022itsa,zhang2022revisiting,rao2023masked,kim2022pointfix,tonioni2019learning}. MS-PSMNet \citep{cai2020matching} replaces learning-based features with hand-crafted ones to force the stereo network to focus on the matching space. Although it achieves significant improvement, its hand-crafted features limit the performance. Thus, many methods improve training via transfer learning \citep{liu2022graftnet}, meta-learning \citep{tonioni2019learning,kim2022pointfix}, contrastive learning \citep{zhang2022revisiting,rao2023masked}, and Fisher information \citep{chuah2022itsa}.

\begin{figure*}[htbp]
\centering
    \includegraphics[width=1.\textwidth]{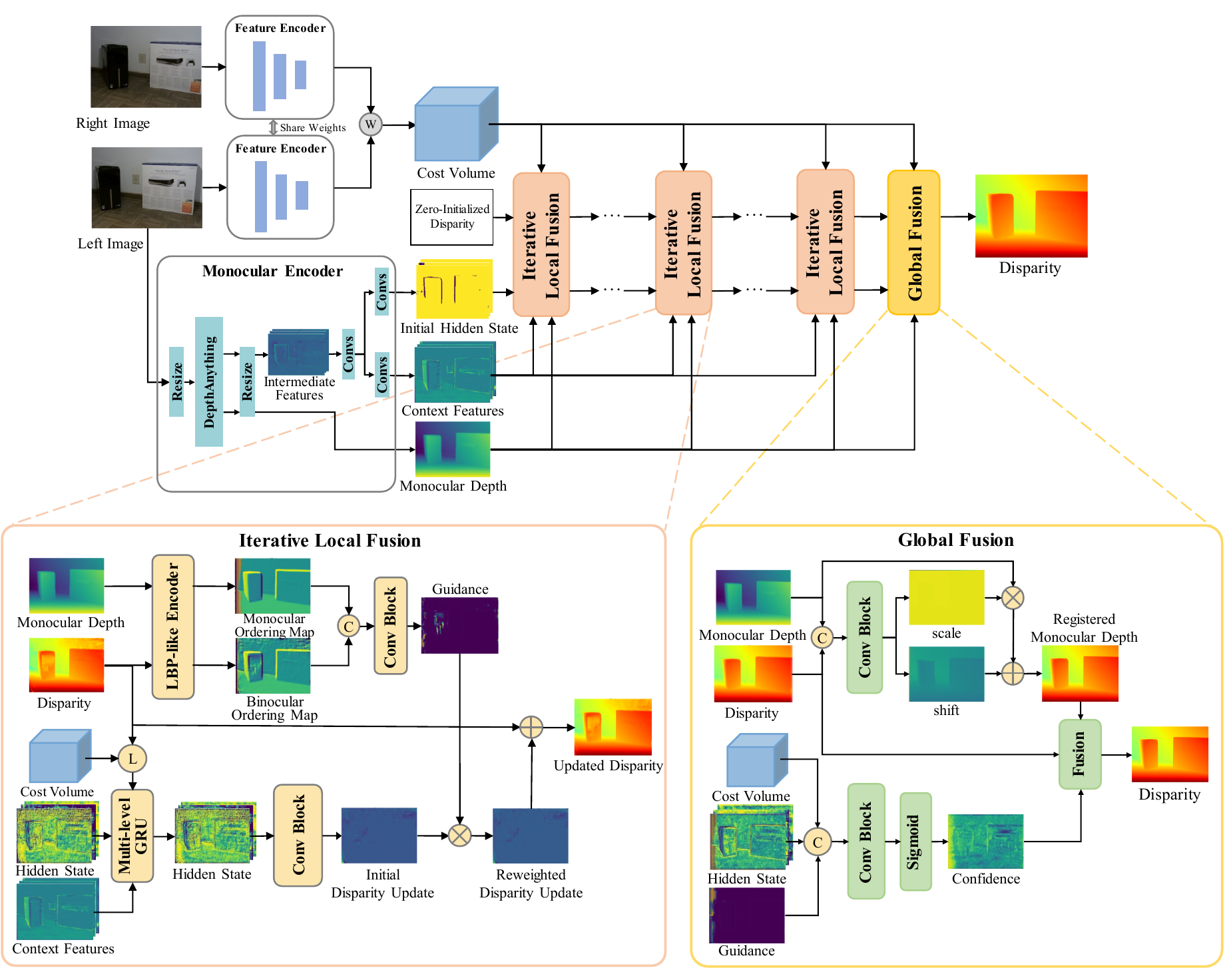}
    % \vspace{-0.55cm}
    \caption{The pipeline of our method. \textcircled{w} represents the warping operation when constructing the cost volume. \raisebox{-.2ex}{\textcircled{\scriptsize L}} is look up operation used to sample cost volume. \textcircled{c} represents concatenation. \textcircled{+} represents add operation, while \textcircled{$\times$} represents multiplication.
}
    \label{Fig: pipeline}
\vspace{-0.5cm}
\end{figure*}

% The above feature-based methods significantly improve the generalization of stereo matching. However, it is difficult for them to eliminate domain gaps due to the complexity of real-world scenarios. Then, some researchers integrate other modalities to enrich the features of RGB images \citep{zhang2022stereo,walz2023gated}. They achieve impressive performance but require additional devices. Instead, other researchers propose to generate more and better data for training \citep{song2022adastereo,chang2023domain,tosi2023nerf, bartolomei2024stereo, wen2025foundationstereo}.
% AdaStereo \citep{song2022adastereo} and HVT-RAFT \citep{chang2023domain} augment the training data in color space to enrich the domain distribution in the synthetic dataset. They improve greatly, but the rendered images in the synthetic dataset are unrealistic to the real world. Thus, NerfStereo \citep{tosi2023nerf} turns to reconstruct the real-world scenes from Nerf and re-renders stereo images to improve the quality of training data. 

The above feature-based methods significantly improve stereo matching generalization. However, due to real-world complexity, they struggle to eliminate domain gaps. Some researchers integrate other modalities to enrich RGB features \citep{zhang2022stereo,walz2023gated}, achieving strong performance but requiring extra devices. Others instead generate more and better training data \citep{song2022adastereo,chang2023domain,tosi2023nerf,bartolomei2024stereo,wen2025foundationstereo}. AdaStereo \citep{song2022adastereo} and HVT-RAFT \citep{chang2023domain} augment data in color space to enrich domain distribution. While effective, rendered images remain unrealistic. Thus, NerfStereo \citep{tosi2023nerf} reconstructs real scenes via NeRF and re-renders stereo images to improve training quality.

% In addition to improving generalization from features and data, some methods focus on architecture design with specific knowledge of stereo matching \citep{zhang2020domain,li2021revisiting,guo2022context,xu2023iterative,chen2024mocha,wang2024selective,guan2024neural}. DSMNet \citep{zhang2020domain} uses long-range matching information in cost aggregation to correct the mismatched points. The improvement in cost aggregation is remarkable, but the additional operations in 3D space are time-consuming. Many methods then turn to incorporating global information when constructing cost volume. STTR \citep{li2021revisiting} and CSTR \citep{guo2022context} use transformers to capture long-range matching information. Other methods \citep{wang2024selective, xu2023iterative, chen2024mocha} build auxiliary volume to augment the original cost volume. There are also some methods resolving the generalization problem from uncertainty learning \citep{jing2023uncertainty,lou2023elfnet}. 

Beyond improving generalization through features and data, some methods focus on architecture design leveraging stereo-specific knowledge \citep{zhang2020domain,li2021revisiting,guo2022context,xu2023iterative,chen2024mocha,wang2024selective,guan2024neural}. DSMNet \citep{zhang2020domain} uses long-range matching in cost aggregation to correct mismatches. While effective, its 3D operations are time-consuming. Many later methods incorporate global information in cost volume construction. STTR \citep{li2021revisiting} and CSTR \citep{guo2022context} apply transformers to capture long-range dependencies. Others \citep{wang2024selective,xu2023iterative,chen2024mocha} build auxiliary volumes to enhance the original cost volume. Some approaches also improve generalization via uncertainty learning \citep{jing2023uncertainty,lou2023elfnet}.

The aforementioned approaches have achieved great performance but still rely on biased monocular priors. Our method introduces unbiased monocular priors from a pre-trained large model and uses effective fusion mechanisms to fuse them, achieving impressive generalization ability.

\subsection{Fusing Monocular and Stereo Estimation}

Inspired by the human visual system fusing binocular disparity and monocular cues \cite{renner2013perception,cutting1995perceiving,welchman2016human,welchman20053d,cheng2025monster,bartolomei2024stereo,wen2025foundationstereo}, researchers have explored similar fusion mechanisms in machine vision. Traditional methods \cite{saxena2007depth} use MRF optimization based on disparity and monocular cues. Deep learning methods mainly adopt volume or depth map fusion \cite{cheng2025monster,bartolomei2024stereo,wen2025foundationstereo}. Volume fusion injects monocular priors into cost volumes \cite{yu2023multi,li2023learning,cheng2025monster,bartolomei2024stereo,wen2025foundationstereo}, but relies on fixed disparity ranges and domain-biased priors. In contrast, our method removes disparity range limits and uses unbiased priors from a large pre-trained model. Depth map fusion methods \cite{martins2018fusion,chen2021revealing,bae2022multi,zhou2023two,aleotti2020reversing} combine monocular and binocular results in post-processing, but often suffer from misalignment and noise due to affine-invariant monocular predictions. Instead, we employ local ordering maps for better compatibility, reducing noise during matching. The final monocular depth is globally aligned to the optimized disparity by learning two parameters to address scale ambiguity.

%% file: sec/3_method.tex
% \begin{figure*}
%     \centering
%     % 第二行下半栏的布局
%     \begin{minipage}[t]{\textwidth}
%         % 左下角的第三张图
%         \begin{minipage}[t]{0.55\textwidth}
%             \centering
%             \includegraphics[width=\textwidth]{Figure/LocalFusion2.pdf}
%             \caption{The architecture of iterative local fusion module. \raisebox{-.2ex}{\textcircled{\scriptsize L}} is look up operation used to sample cost volume. \textcircled{c} represents concatenation. \textcircled{+} represents add operation, while \textcircled{$\times$} represents multiplication.}
%             \label{Fig: local fusion}
%         \end{minipage}
%         \hfill
%         % 右下角的第四张图
%         \begin{minipage}[t]{0.425\textwidth}
%             \centering
%             \includegraphics[width=\textwidth]{Figure/GlobalFusion2.pdf}
%             \caption{The architecture of global fusion module. \textcircled{c} represents concatenation. \textcircled{+} represents add operation, while \textcircled{$\times$} represents multiplication.}
%             \label{Fig: global fusion}
%         \end{minipage}
%     \end{minipage}
%     \vspace{-0.3cm}
% \end{figure*}

\section{Method}
\label{sec:method}
Our network structure is illustrated in Figure \ref{Fig: pipeline}. First, we extract features from the left and right images to construct a cost volume. Meanwhile, the monocular encoder module extracts initial hidden states, context features, and monocular depth from the left image using a pre-trained large monocular model \cite{yang2024depthv2}. Then, the local fusion module iteratively optimizes the disparity estimation with monocular priors using the local binary ordering map. Finally, the global fusion module registers the optimized disparity with the monocular depth as the final result.

\subsection{Monocular Encoder}
The monocular priors learned by the stereo-matching model are heavily biased due to the scarcity of wild-world stereo data \citep{guo2017calibration,ovadia2019can}. This paper uses the widely used DepthAnything v2\cite{yang2024depthv2}  to extract unbiased monocular priors to mitigate the domain gap, including monocular context features and depth. However, it is flexible to use other VTFs as long as the monocular prior is not biased to specific scenarios.

As shown in Figure \ref{Fig: pipeline}, given an image with a resolution of $H \times W$, we pre-process the image as DepthAnything v2 \cite{yang2024depthv2} by resizing the longest side of the image to 512 pixels. The resized image is then fed into a frozen DepthAnything v2 to extract intermediate features before the DPTHead \cite{ranftl2021vision} and monocular depth after the DPTHead. These intermediate features and monocular depth are subsequently resized to a $H/4 \times W/4$ resolution using a bilinear function to interact with the stereo-matching pipeline. We build a two-stream convolution module to generate the initial hidden state and monocular context features from the intermediate features. Although both the prediction and supervision of DepthAnything v2 are in the form of inverse depth, which is disparity under some unknown camera parameters and baseline, we still refer to it as monocular depth here to maintain consistency with the terminology used in DepthAnything v2.
% It is worth noting that the monocular depth output from DepthAnything v2 is the inversed version of depth, which can be deemed as a disparity. To maintain consistency with the terminology used in DepthAnything v2, we still refer to it as monocular depth here.

\subsection{Iterative Local Fusion}

The iterative local fusion module leverages the binary local ordering map to update disparity with monocular priors iteratively.
The binary ordering map \(M_O\) encodes the relative ordering of depth $D$ between a center pixel and its neighbors: 
\begin{equation}
    M_O(u,v) = \{ \sigma(D(u',v') - D(u, v)) \}, 
\end{equation}
where $(u',v') \in \mathcal{N}_{(u,v)}$, \(\sigma\) is the sigmoid function and \(\mathcal{N}\) is the neighborhood. 
The binary local ordering map helps mitigate the impact of outlier noises by converting absolute values into ordering relationships, which is much more robust than the pixel-wise depth value. Besides, it also unifies the affine-invariant depth and absolute disparity to be compatible with the order relationship. 

To compute the binary local ordering map, we use a series of LBP-like operations \cite{ojala2002multiresolution,ojala1994performance} with varying window sizes to extract local ordering features. Each LBP-like operation consists of a convolution with fixed weights followed by a sigmoid function, which measures the relative depth relationships between the center pixel of the window and its neighboring pixels, indicating which pixels are closer or farther. To employ the binary local ordering map into the iterative refinement structure, we use the LBP-like encoder to extract local ordering maps from both monocular depth and binocular disparity in the previous iteration, as shown in Figure \ref{Fig: pipeline}. These two kinds of local ordering maps are concatenated to predict the monocular guidance. 
The guidance \(G\) is modeled by a Beta distribution with parameters \( \{\alpha, \beta\} \) predicted via convolutions. During training, \(G\) is sampled via reparameterization: \(G = g_1 / (g_1 + g_2)\), \(g_1 \sim \text{Gamma}(\alpha, 1)\), \(g_2 \sim \text{Gamma}(\beta, 1)\). At test time, \(G = \alpha / (\alpha + \beta)\). 
The guidance $G$ is then used to re-weight the initial disparity update $\Delta_d$ to avoid local optima. 
% We represent the monocular guidance as a Beta distribution with a manually specified amplitude parameter. During training, we apply the re-parameterization trick to obtain guidance, while during testing, we compute the distribution expectation to obtain it. 

As we mentioned, the first several disparity predictions are noisy, especially during training. The local ordering map may still have many wrong relative depth values, leading to wrong guidance and slow training convergence. Therefore, we propose to gradually release the influence of guidance to the initial disparity update results as
\begin{equation} 
    \Tilde{\Delta}_d = \Delta_d (1 + G \cdot r \cdot t/T). 
\end{equation} 
Here, $r$ is the manually specified amplitude parameter that controls the influence of the guidance.  $t$ represents the current iteration number, and $T$ is the total number of iterations. The initial disparity update $\Delta_d$ is predicted by a multi-level GRU followed by a convolution block. Finally, the disparity is updated by adding the re-weighted disparity update to the disparity from the previous iteration: 
\begin{equation} 
    D_d^{t} = D_d^{t-1} + \Tilde{\Delta}_d. 
\end{equation}

\subsection{Global Fusion}
After all iterations of disparity update, we use a global fusion module to incorporate fine-grained 3D shape priors from the monocular depth map into the disparity map, as shown in Figure \ref{Fig: pipeline}. Here, we formulate the optimized binocular disparity as a registered version of monocular depth with minor noise by specific intrinsic parameters. Therefore, monocular depth can be globally registered to binocular disparity. The registration can be deemed a linear regression problem with noise between monocular depth and binocular disparity. To this end, we first align the monocular depth $D_m$ with the optimized disparity $D_d$ by estimating two registration parameters, ${a, b}$ by
\begin{equation}
    \begin{aligned} 
        \Tilde{D}_m &= a \cdot D_m + b, \\
        {a, b} &= \mathcal{F}(D_m, D_d^T), 
    \end{aligned} 
\end{equation}
where $\mathcal{F}$ represents a network with a series of convolution layers and ReLU activation, which take the concatenation of the monocular depth $D_m$ and the optimized disparity $D_d^T$ as input. Simultaneously, we use the sampled cost volume, hidden state, and weights from the previous iteration to predict a confidence map. This confidence, $c$, is then used to fuse the aligned monocular depth $\Tilde{D}_m$ and the optimized disparity $D_d$ as follows 
\begin{equation}
    D_f = c \cdot D_d^T + (1-c) \cdot \Tilde{D}_m. 
\end{equation}
$D_f$ is the final disparity prediction.

\subsection{Loss}
We use $L_1$ loss to supervised the learning of each updated disparity $D_d^t$, registered monocular depth $\Tilde{D}_m$, and the final output of our method $D_f$:
\begin{equation}
\begin{aligned}
% \resizebox{.5\textwidth}{!}{$
%     \mathcal{L} = \sum_{t=1}^T \gamma^{T+2-t} || D_d^t - D_G ||_1 + \gamma || \Tilde{D}_m - D_G ||_1 + || D_f - D_G ||_1.
% $}
    \mathcal{L} = &\sum_{t=1}^T \gamma^{T+2-t} || D_d^t - D_G ||_1 \\
                  & + \gamma || \Tilde{D}_m - D_G ||_1 + || D_f - D_G ||_1.
\end{aligned}
\end{equation}
$D_G$ is the ground-truth disparity. $\gamma$ is the balancing scalar.

% \begin{figure}
% \centering
%     \includegraphics[width=.45\textwidth]{Figure/Encoder.pdf}
%     % \vspace{-0.55cm}
%     \caption{The architecture of our monocular encoder module.}
%     \label{Fig: MGR}
% \end{figure}

% \begin{figure*}
% \centering
%     \includegraphics[width=.7\textwidth]{Figure/LocalFusion.pdf}
%     % \vspace{-0.55cm}
%     \caption{The architecture of our iterative local fusion module.}
%     \label{Fig: MGR}
% \end{figure*}

% \begin{figure*}
% \centering
%     \includegraphics[width=.5\textwidth]{Figure/GlobalFusion.pdf}
%     % \vspace{-0.55cm}
%     \caption{The architecture of our global fusion module.}
%     \label{Fig: MGR}
% \end{figure*}

%% file: sec/4_experiments.tex
\section{Experiments}
\label{sec:experiments}

\subsection{Implementation Details}
For the stereo part, our pipeline is built on the classical iterative structure of RAFT-Stereo \cite{lipson2021raft}, which is widely used and flexible to deploy without stacking network tricks to raise the computation burden. As for the monocular part, we use DepthAnything V2 \cite{yang2024depth,yang2024depthv2} to extract unbiased monocular priors. Still, it is flexible to use other VTFs as long as the monocular prior is generalizable to practical scenarios. We set parameters as $r=1$ and $\gamma=0.9$. The window sizes for the LBP-like operations are configured to ${5, 3}$. The training was conducted on 4 NVIDIA A40 GPUs using the AdamW optimizer with a one-cycle learning rate schedule. During training, the DepthAnything V2 module remains frozen. Specifically, we first train the model without the global fusion module on the SceneFlow dataset, using a maximum learning rate of 0.0002, a batch size of 8, and for 100k steps, maintaining the consistency of matching parts with the total data used in RAFT-Stereo. Then, we train the monocular registration of the global fusion module while keeping the other modules frozen, using a maximum learning rate of 0.0005 and a batch size of 32 for 100k steps on the SceneFlow dataset. Finally, we train the entire global fusion module while keeping the other modules frozen, using a maximum learning rate of 0.0005, a batch size of 32, and 100k steps on the SceneFlow dataset. Our results are not sensitive to the hyperparameters of the training process. 
% With the training and testing codes provided in \hyperlink{https://github.com/YaoChengTang/Diving-into-the-Fusion-of-Monocular-Priors-for-Generalized-Stereo-Matching}{our project}, All the evaluation results can simply be reproduced.

% Table generated by Excel2LaTeX from sheet 'Sheet4'
\begin{table*}[htbp]
  \centering
  \scalebox{0.9}{
  \renewcommand\arraystretch{1.}
  \setlength{\tabcolsep}{1.2mm}
    \begin{tabular}{c|c|c|cc|cc|cc|cc|cc|cc}
    \hline
    \multirow{3}[3]{*}{Method} & \multirow{3}[3]{*}{Year} & \multirow{3}[3]{*}{ \tabincell{c}{Additional \\ Data/Aug} } & \multicolumn{2}{c|}{KITTI 2015} & \multicolumn{2}{c|}{KITTI 2012} & \multicolumn{6}{c|}{Middlebury (H)} & \multicolumn{2}{c}{ETH3D} \\
        \cline{4-13} &         &       & \multirow{2}[2]{*}{EPE} & \multirow{2}[2]{*}{bad 3.0} & \multirow{2}[2]{*}{EPE} & \multirow{2}[2]{*}{bad 3.0} & \multicolumn{2}{c|}{All} & \multicolumn{2}{c|}{NonOcc} & \multicolumn{2}{c|}{Occ} & \multirow{2}[2]{*}{EPE} & \multirow{2}[2]{*}{bad 1.0} \\
            &      &       &       &       &       &       & \multicolumn{1}{c}{EPE} & \multicolumn{1}{c|}{bad 2.0} & \multicolumn{1}{c}{EPE} & \multicolumn{1}{c|}{bad 2.0} & \multicolumn{1}{c}{EPE} & \multicolumn{1}{c|}{bad 2.0} &       &  \\
    \hline
    FC-PSMNet \citep{zhang2022revisiting} & 2022 &      & 1.58 & 7.50  & 1.42 & 7 & 4.14 & 18.3 & - & - & - & - & 1.25 & 12.8 \\
    ITSA-PSMNet \citep{chuah2022itsa} & 2022 &      & 1.39 & 5.80  & 1.09 & 5.2 & 3.25 & 12.7 & - & - & - & - & 0.94 & 9.8 \\
    Graft-PSMNet \citep{liu2022graftnet} & 2022 &      & 1.32 & 5.30  & 1.09 & 5 & 2.34 & 10.9 & - & - & - & - & 1.16 & 10.7 \\
    Mask-CFNet \citep{rao2023masked} & 2023 &      & - & 5.80  & - & 4.8 & - & 13.7 & - & - & - & - & - & 5.7 \\
    STTR* \citep{li2021revisiting} & 2021 &      & 2.14 & 9.5  & 2.51 & 9.62 & 9.13 & 21.76 & 5.03 & 13.49  & 35.98 & 78.84 & - & - \\
    PCWNet \citep{shen2022pcw} & 2022 &      & - & 5.60  & - & 4.2 & - & 15.8 & - & 15.8 & - & - & 3.8 & 14.4 \\
    RAFTStereo* \citep{lipson2021raft} & 2021 &      & 1.13 & 5.69  & 0.9 & 4.35 & 1.92 & 12.6 & 1.09 & 8.65 & 3.31 & 26.39 & 0.36 & 3.3 \\
    % CREStereo \citep{li2022practical} & 2022 &      & - & 6.70  & - & 6.1 & - & 15.3 &   -    &  -     & - & 5.5 \\
    IGEV* \citep{xu2023iterative} & 2023 &      & 1.21 & 6.03  & 1.03 & 5.13 & 2.63 & 11.93 & 2.27 & 9.49 & 5.02 & 26.04 & 0.33 & 4 \\
    ELFNet* \citep{lou2023elfnet} & 2023 &      & 2.31 & 7.68  & 1.36 & 5.85 & 5.16 & 17.5 & 2.16 & 10.14 & - & - &  - & - \\
    Mocha-Stereo* \cite{chen2024mocha} & 2024 &     & 1.29 & 5.97 & 1.02 & 4.83  & 2.66  & 10.18  & 2.49  & 7.96 & 3.84 & \textbf{24.16}  & 0.28 & 3.47 \\
    NMRF* \cite{guan2024neural}  & 2024 &      & 1.17 & 5.31 & 0.92 & 4.63 & 2.91  & 13.36  & 2.73  & 10.90 & - & -  &  0.31 & 3.8 \\
    Selective-RAFT* \cite{wang2024selective} & 2024 &      & 1.27 & 6.68 & 1.08  & 5.19 & 2.34  & 12.04  & 2.05  & 9.45 & 4.17 & 27.4  & 0.34 & 4.36 \\
    Selective-IGEV* \cite{wang2024selective} & 2024 &      & 1.25 & 6.06 & 1.08  & 5.64 & 2.59  & 11.79  & 2.31  & 9.22 & 4.35 & 28.10  & 0.33 & 4.05 \\
    HVT-RAFT \cite{chang2023domain} & 2023 &   \checkmark   & 1.12 & \textbf{5.20}  & 0.87 & 3.7 & 1.37  & 10.40  &   -    &   -  & - & -   & 0.29 & 3.00  \\
    NerfStereo* \cite{tosi2023nerf} & 2023 &  \checkmark   & 1.14 & 5.41  & \textbf{0.84} & \textbf{3.6} & 1.42 & 9.67  & 0.91  & 6.39 & 4.09 & 29.89 & 0.29 & 2.94  \\
    \hline
    RAFT-Stereo + ME &  &      & 1.18  & 6.18  & 0.87  & 4.19  & 1.42  & 9.73  & 1.11  & 7.00 & 3.06 & 26.50 & 0.26  & 2.31  \\
    Ours  &  &      & \textbf{1.12}  & 5.60  & \textbf{0.87}  & 4.10  & \textbf{1.15}  & \textbf{8.39}  & \textbf{0.85}  & \textbf{5.67}  & \textbf{2.89} & 26.50 & \textbf{0.25}  & \textbf{1.88}  \\
    \hline
    \end{tabular}%
    }
    \captionsetup{skip=3pt} % 设置特定表格的标题与正文距离
  \caption{Generalization from SceneFlow dataset to KITTI2015, KITTI 2012, Middlebury (H), and ETH3D dataset. `ME' represents our monocular encoder module. * represents the results evaluated in our metrics and settings using official models and weights. `All', `NonOcc', and `Occ' represent all regions, non-occluded regions, and occluded regions, respectively.}
  \label{tab: mixed datasets}%
  \vspace{-0.2cm}
\end{table*}%

\begin{table*}[htbp]
  \centering
  \scalebox{0.9}{
  \renewcommand\arraystretch{1.}
  \setlength{\tabcolsep}{.65mm}
    \begin{tabular}{c|c|cccc|cccc|cccc}
    \hline
    \multirow{3}[3]{*}{Method} & \multirow{3}[3]{*}{ \tabincell{c}{Additional \\ Data/Aug} } & \multicolumn{12}{c}{Booster (Q)} \\
        \cline{3-14}          &       & \multicolumn{4}{c|}{ALL} & \multicolumn{4}{c|}{Trans} & \multicolumn{4}{c}{NonTrans} \\
                  &       & \multicolumn{1}{c}{EPE} & \multicolumn{1}{c}{bad 2.0} & \multicolumn{1}{c}{bad 3.0} & \multicolumn{1}{c|}{bad 5.0} & \multicolumn{1}{c}{EPE} & \multicolumn{1}{c}{bad 2.0} & \multicolumn{1}{c}{bad 3.0} & \multicolumn{1}{c|}{bad 5.0} & \multicolumn{1}{c}{EPE} & \multicolumn{1}{c}{bad 2.0} & \multicolumn{1}{c}{bad 3.0} & \multicolumn{1}{c}{bad 5.0} \\
    \hline
    Mocha-Stereo \cite{chen2024mocha} &       & 3.88  & 16.82 & 14.31 & 11.84 & 9.45  & 66.44 & 57.96 & 45.73 & 2.89  & 12.31 & 10.19 & 8.38 \\
    ELFNet \cite{lou2023elfnet} &       &6.05       &24.51      &20.43     &16.40      &9.03     &72.07     &62.73    &49.82    &5.33   &20.85    &17.18    &13.84\\
    Selective-RAFT \cite{wang2024selective} &       & 4.14  & 19.52 & 16.69 & 13.63 & 10.34 & 69.84 & 61.64 & 49.55 & 2.99  & 14.99 & 12.44 & 10.00 \\
    Selective-IGEV \cite{wang2024selective} &       & 4.62  & 19.28 & 16.58 & 13.92 & 9.50   & 66.85 & 58.9  & 47.15 & 3.60   & 14.74 & 12.34 & 10.27 \\
    IGEV \cite{xu2023iterative} &       & 4.26 & 17.58 & 15.21 & 12.89 & 10.00 & 68.96 & 61.14 & 49.51 & 3.25 & 12.99 & 10.94 & 9.24 \\
    NMRF \cite{guan2024neural} &       &  5.05 &  26.22 & 21.31 & 16.58 & 10.36 &  70.92 & 60.93 & 47.16 & 4.00 & 22.43 & 17.77 & 13.50 \\
    NerfStereo \cite{tosi2023nerf} & \checkmark & 3.48 & 13.40 & 11.13 & 9.22 & 8.88 & 62.67 & 53.35 & 41.79 & 2.49 & 9.06 & 7.19 & 5.89 \\
    RAFTstereo \cite{lipson2021raft} &       & 4.18 & 17.64 & 14.92 & 12.23 & 9.79 & 67.69 & 59.31 & 47.40 & 3.23 & 13.13 & 10.75 & 8.70 \\
    \hline
    RAFT-Stereo + ME &       & 2.40 & 11.44 & 9.17 & 7.30 & 8.97 & 64.84 & 56.05 & 43.95 & \textbf{1.45} & \textbf{6.96} & 5.08 & 3.89 \\
    \textbf{Ours} &       & \textbf{2.26} & \textbf{11.02} & \textbf{8.59} & \textbf{6.6} & \textbf{7.93} & \textbf{59.83} & \textbf{50.36} & \textbf{38.44} & 1.52 & 6.98 & \textbf{4.97} & \textbf{3.64} \\
    \hline
    \end{tabular}%
    }
    \captionsetup{skip=3pt} % 设置特定表格的标题与正文距离
  \caption{Generalization from SceneFlow dataset to Booster dataset in quarter resolution and balanced set. `ME' represents our monocular encoder module. `All', `Trans', and `NonTrans' represent all regions, transparent regions, and nontransparent regions, respectively.}
  \vspace{-0.4cm}
  \label{tab: booster}%
\end{table*}%

% \begin{figure*}[htbp]
% \centering
%     \includegraphics[width=1\textwidth]{Figure/fig1-all.pdf}
%     % \vspace{-0.55cm}
%     \caption{The visualization of different ill-posed areas in the Booster dataset.}
%     \vspace{-0.3cm}
%     \label{Fig: vis booster}
% \end{figure*}

\subsection{Evaluation}
\noindent\textbf{Datasets.} Domain generalized stereo matching is typically trained on the SceneFlow dataset \cite{mayer2016large} and evaluated on the training sets of various real-world datasets. We select five real-world datasets, each containing different ill-posed regions, to evaluate the in-the-wild generalization ability of the models, including KITTI 2012 \cite{Geiger2012CVPR}, KITTI 2015 \cite{Menze2018JPRS, Menze2015ISA}, Middlebury \cite{scharstein2014high}, ETH3D \cite{schops2017multi}, and Booster \cite{ramirez2022open}.

\noindent\textbf{Metrics.} (1) We use two metrics: EPE, which measures the mean absolute disparity error in pixels, and Bad $x$, which represents the percentage of pixels where the predicted disparity deviates from the ground truth by at least $x$ pixels. (2) It is important to note that many recent methods report their results with some implicit assumptions, such as evaluating only pixels with ground truth disparity less than 192 or only evaluating non-occluded regions. In our experiments, unless otherwise specified, both ours and the compared methods consider all regions as the classical metric does without limitations. For the Middlebury dataset, we evaluate both all regions and non-occluded regions. For the Booster dataset, we evaluate all regions, as well as transparent and non-transparent regions. (3) Additionally, we observe fluctuations in model performance when trained with different numbers of steps. To fully analyze the improvement contributed by each model component, we calculate the mean and standard deviation (std) of results from the last 100k, 90k, and 80k training steps. We use $mean \pm std$ to measure the accuracy and robustness of our model.

\subsection{In-the-wild Generalization Ability}
\begin{table}[htbp]
  \centering
  \scalebox{0.9}{
  \renewcommand\arraystretch{1.}
  \setlength{\tabcolsep}{1.mm}
    \begin{tabular}{c|c|cccc}
    \hline
    \multirow{1}[4]{*}{Method} & \multirow{1}[4]{*}{ \tabincell{c}{Additional \\ Data/Aug} } & \multicolumn{4}{c}{DrivingStereo} \\
\cline{3-6}          &       & Sunny & Cloudy & Rainy & Foggy \\
    % Method  & \tabincell{c}{Additional \\ Data/Aug} & Sunny & Cloudy & Rainy & Foggy \\
    \hline
    RAFTStereo \cite{lipson2021raft} &       & 1.01  & 0.97  & 1.8   & 0.95 \\
    IGEV \cite{xu2023iterative} &       & 1.11  & 1.11  & 2.32  & 1.14 \\
    MoCha-Stereo \cite{chen2024mocha} &       & 1.01  & 0.99  & 1.35  & 0.98 \\
    Selective-IGEV \cite{wang2024selective} &       & 1.18  & 1.13  & 2.22  & 1.12 \\
    NerfStereo \cite{tosi2023nerf} &   \checkmark   & \textbf{0.90} & \textbf{0.91} & 1.46  & 1.01 \\
    \hline
    Ours  &       & \textbf{0.93} & \textbf{0.92} & \textbf{1.29} & \textbf{0.93} \\
    \hline
    \end{tabular}%
  }
  \label{Table: Driving}%
  \vspace{-1mm}
  \caption{Generalization from SceneFlow to DrivingStereo. EPE is used as the evaluation metric.}
  % \vspace{-8.3mm}
\end{table}%
As shown in Table \ref{tab: mixed datasets}, our method achieves state-of-the-art results across all datasets, with particularly strong performance on Middlebury and ETH3D. Compared to other methods that do not use additional data or augmentation, we almost double their performance. Even when compared to methods incorporating additional data or augmentation, our approach leverages limited stereo data to achieve superior results. Furthermore, as presented in Table \ref{tab: booster}, our method demonstrates substantial improvements in the Booster dataset. Compared to methods without additional data or augmentation, we nearly double the improvement on EPE and Bad 5.0 across all regions, achieve more than a 10-point improvement on Bad x.0 in transparent regions, and show double or even triple the improvement in non-transparent regions. For more detailed quantitative results and analysis, please refer to our supplementary materials.

\begin{table}[htbp]
  \centering
    \renewcommand\arraystretch{.9}
    % 子表格一
    \begin{subtable}[t]{.5\textwidth}
      \centering
      \setlength{\tabcolsep}{1mm}
      \scalebox{0.75}{
      \renewcommand\arraystretch{1.1}
      \begin{tabular}{c|cc|cc|cc}
        \hline
        \multirow{1}[4]{*}{Method} & \multicolumn{2}{c|}{All} & \multicolumn{2}{c|}{NonOcc} & \multicolumn{2}{c}{Occ} \\
        \cline{2-7}
        & EPE $\downarrow$  & bad 2.0 $\downarrow$  & EPE $\downarrow$  & bad 2.0 $\downarrow$ & EPE $\downarrow$  & bad 2.0 $\downarrow$ \\
        \hline
        DA V2 - M &  205.04 &  99.99 &  207.51 &  99.99 &  196.96 & 99.98 \\
        \hline
        DA V2 - GA &  5.83 &  69.28 &  5.61 &  69.16 &  6.95 & 69.34 \\
        \hline
        Metric3D &  33.14 &  97.18 &  33.05 &  97.06 &  34.54 & 98.09 \\
        \hline
        Ours  & \textbf{1.15} & \textbf{8.39} & \textbf{0.85} & \textbf{5.67} & \textbf{2.89} & \textbf{26.50} \\
        \hline
      \end{tabular}
      }
      % \vspace{-1mm}
      \caption{Metric disparity space}
    \end{subtable}
  % \vspace{3mm} % 控制子表格间距
  
    % 子表格二
    \begin{subtable}[t]{.5\textwidth}
      \centering
      \setlength{\tabcolsep}{2mm}
      \scalebox{0.75}{
      \renewcommand\arraystretch{1.1}
      \begin{tabular}{c|cc|cc|cc}
        \hline
        \multirow{1}[4]{*}{Method} & \multicolumn{2}{c|}{All} & \multicolumn{2}{c|}{NonOcc} & \multicolumn{2}{c}{Occ} \\
        \cline{2-7}
        & $\delta^1$ $\uparrow$ & RMS $\downarrow$ & $\delta^1$ $\uparrow$ & RMS $\downarrow$ & $\delta^1$ $\uparrow$ & RMS $\downarrow$ \\
        \hline
        DA V2 - M &  0.022 &  6.356 &  0.024 &  6.171 &  0.008 &  7.189 \\
        \hline
        DA V2 - GA &  0.923 &  1.487 &  0.934 &  1.342 &  0.852 &  2.022 \\
        \hline
        Metric3D &  0.288 & 4.097 &  0.298 & 4.014 &  0.223 &  4.588 \\
        \hline
        Ours  & \textbf{0.985} & \textbf{0.677} & \textbf{0.991} & \textbf{0.492} & \textbf{0.948} & \textbf{1.282} \\
        \hline
      \end{tabular}
      }
      % \vspace{-1mm}
      \caption{Metric depth space}
    \end{subtable}
    \vspace{-4mm}
  \caption{Comparison to DepthAnything V2 and Metric3D on Middlebury (H). M: the fine-tuned metric version. GA: alignment to GT scale using identical registration parameters computed from GT for all pixels.}
  \label{Table: DA}
  \vspace{-4mm}
\end{table}

\begin{table}[htbp]
  \centering
  \scalebox{0.95}{
  \renewcommand\arraystretch{1.}
  \setlength{\tabcolsep}{1.8mm}
    \begin{tabular}{c|c|c}
    \hline
    \multirow{2}[2]{*}{Exp} & \multicolumn{2}{c}{Middlebury (H)} \\
        \cline{2-3}          & epe   & bad 2.0 \\
    \hline
    Baseline & 2.11$\pm$0.16 & 14.12$\pm$0.64 \\
    Baseline w/o mono feature & 1.83$\pm$0.11 & 12.45$\pm$0.86 \\
    Baseline + ME & 1.42$\pm$0.01 & 9.81$\pm$0.18 \\
    Baseline + ME + IDF & 1.41$\pm$0.04 & 10.34$\pm$0.19 \\
    Baseline + ME + PF  & 1.41$\pm$0.00 & 9.71$\pm$0.00 \\
    Baseline + ME + ILF & 1.20$\pm$0.08 & 9.06$\pm$0.70 \\
    Baseline + ME + ILF + GF & 1.15$\pm$0.01 & 8.35$\pm$0.04 \\
    \hline
    \end{tabular}%
    }
  \captionsetup{skip=3pt} % 设置特定表格的标题与正文距离
  \caption{The effectiveness of each module. The baseline is RAFTStereo. w/o mono feature: removing the context features from RAFTStereo. ME:  our monocular encoder, DF: iterative direct fusion, ILF: iterative local fusion, PF: post-fusion, GF: global fusion.}
  \label{tab: abl module}%
  \vspace{-0.5cm}
\end{table}%

\begin{figure}[htbp]
\centering
    \includegraphics[width=.43\textwidth]{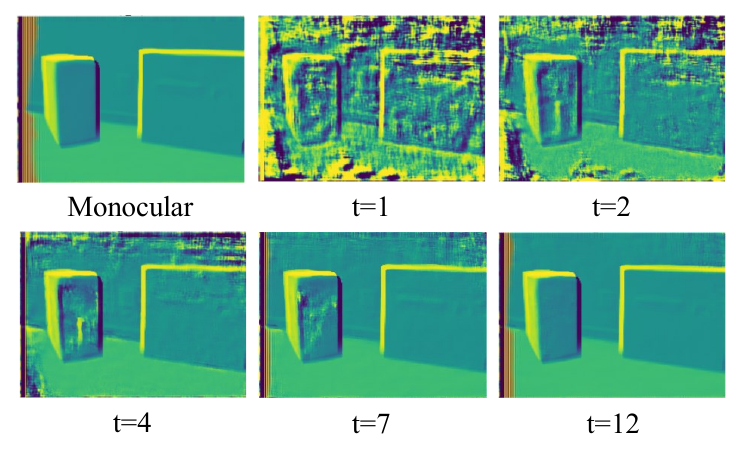}
    \vspace{-0.4cm}
    \caption{The visualization of local ordering map. The monocular represents the results from monocular depth. $t=x$ represents the results from binocular disparity.}
    \label{Fig: vis lbp}
    \vspace{-0.5cm}
\end{figure}

\begin{table}[htbp]
  \centering
  \scalebox{0.9}{
  \renewcommand\arraystretch{1.}
  \setlength{\tabcolsep}{.8mm}
    \begin{tabular}{ccccc|c|c}
    \hline
    \multirow{2}[2]{*}{Exp} & \multirow{2}[1]{*}{ \tabincell{c}{LBP\\Kernel} } & \multirow{2}[2]{*}{S} & \multirow{2}[2]{*}{OP} & \multirow{2}[2]{*}{ r } & \multicolumn{2}{c}{Middlebury (H)} \\
\cline{6-7}          &       &       &       &       & \multicolumn{1}{c|}{epe} & bad 2.0 \\
    \hline
    % DirectFusion &       &       &       & 1     & 1.41$\pm$0.04 & 10.34$\pm$0.19 \\
    L(1)  & 1     &       & L   & 1     & 1.38$\pm$0.06 & 10.20$\pm$0.53 \\
    L(2)  & 3     &       & L   & 1     & 1.36$\pm$0.04 & 10.10$\pm$0.25 \\
    L(3)  & 3     & \checkmark   & L   & 1     & 1.44$\pm$0.06 & 9.65$\pm$0.26 \\
    L(4)  & 5,3   & \checkmark   & L   & 1     & 1.20$\pm$0.08 & 9.06$\pm$0.70 \\
    L(5)  & 9,7,5,3 & \checkmark   & L   & 1     & 1.32$\pm$0.07 & 9.57$\pm$0.62 \\
    L(6)  &       &   \checkmark    & C  & 1     & 1.32$\pm$0.14 & 9.53$\pm$1.03 \\
    L(7)  &       &   \checkmark    & DC  & 1     & 1.39$\pm$0.03 & 9.71$\pm$0.29 \\
    L(8)  & 13,11,9,7,5,3 &   \checkmark    & L   & 1     & 1.31$\pm$0.02 & 9.89$\pm$0.04 \\
    L(9)  & 5,3,  &   \checkmark    & L   & 2     & 1.32$\pm$0.03 & 9.45$\pm$0.14 \\
    L(10) & 5,3   &   \checkmark    & L   & 3     & 1.26$\pm$0.09 & 9.42$\pm$0.42 \\
    \hline
    \end{tabular}%
}
\vspace{-0.2cm}
  \caption{Ablation study on iterative local fusion. S: an LBP-like operation with or without a sigmoid function. OP: the type of operation, L: the LBP-like operation, C: the convolution, DC: a deeper convolution, r: the amplitude parameter.}
  \label{tab: ILF}%
  \vspace{-0.1cm}
\end{table}%

\begin{table}[htbp]
  \centering
  \scalebox{0.9}{
  \renewcommand\arraystretch{1.1}
  \setlength{\tabcolsep}{.8mm}
    \begin{tabular}{ccc|c|c}
    \hline
    \multirow{2}[2]{*}{Exp} & \multirow{2}[2]{*}{Reg} & \multirow{2}[2]{*}{Confidence} & \multicolumn{2}{c}{Middlebury (H)} \\
        \cline{4-5}          &       &       & epe   & bad 2.0 \\
    \hline
    % OnlyPostFusion &       & Cost  & 1.41$\pm$0.00 & 9.71$\pm$0.00 \\
    G(1)  &       & Cost  & 1.23$\pm$0.03 & 9.69$\pm$0.38 \\
    MonoDepth  & \checkmark   &   & 1.19$\pm$0.02 & 8.72$\pm$0.20 \\
    G(2)  & \checkmark   & Cost  & 1.18$\pm$0.02 & 8.77$\pm$0.07 \\
    G(3)  & \checkmark   & Hybrid & 1.15$\pm$0.01 & 8.35$\pm$0.04 \\
    \hline
    \end{tabular}%
    }
    \vspace{-0.2cm}
  \caption{Ablation study on the global fusion. Reg: registration for monocular depth. Cost: estimating the confidence from the sampled cost volume. Hybrid: estimating the confidence from the concatenation of sampled cost volume, hidden state, and guidance from the last iteration. MonoDepth: evaluation of the registered monocular depth.}
  \label{tab: GF}%
  \vspace{-0.4cm}
\end{table}%

We also provide visualization results on the Booster dataset to show the zero-shot generalization ability of our method in the wild world. As illustrated in Figure \ref{fig: fig1}, our method significantly improves performance in various challenging regions, such as areas with occlusion, textureless surfaces, reflections, and transparent regions. Due to space limitations, additional visualization results are available in the supplementary materials.

\begin{figure*}[htbp]
\vspace{-0.3cm}
\centering
    \includegraphics[width=.95\textwidth]{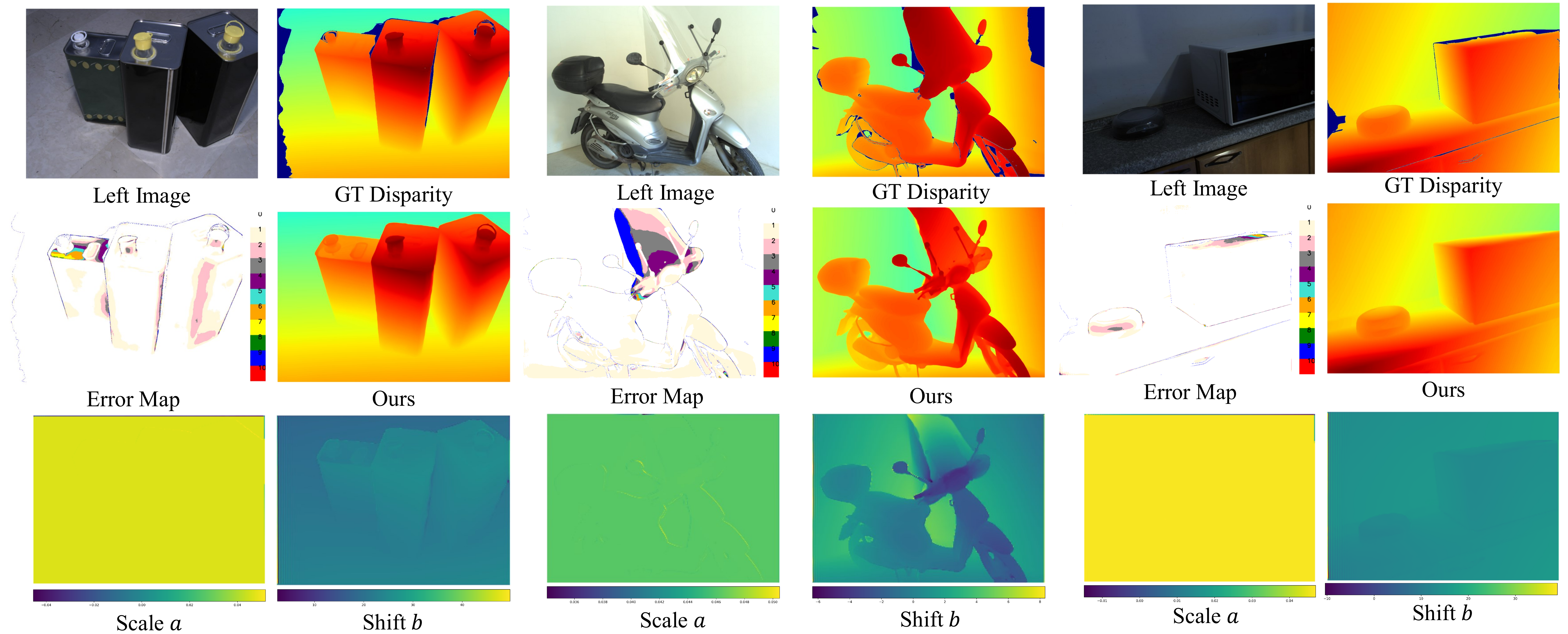}
    \vspace{-0.1cm}
    \caption{The visualization of registration parameters, scale $a$ and shift $b$.}
    \label{Fig: vis reg}
    \vspace{-0.4cm}
\end{figure*}

\subsection{Ablation Study and Analysis}
We conduct comprehensive ablation studies to analyze the impact of each module and illustrate the construction process of our model. It is important to note that each ablation study involves training the model from scratch rather than removing a component from an already well-trained model.

\noindent\textbf{The Effectiveness of Each Module.}
As shown in Table \ref{tab: abl module}, the baseline model performs better without context features, indicating that monocular priors are susceptible to domain bias when data is limited. By incorporating less-biased monocular priors from a pre-trained large monocular network in the monocular encoder (ME), generalization performance is significantly improved, highlighting the importance of robust monocular priors in the wild world. Comparing the Baseline + ME with Baseline in Table \ref{tab: mixed datasets}, its performance becomes worse than Raft-Stereo, showing that it is easy to suffer from over-confidence when simply fusing monocular features with disparities during iterative disparity update. The iterative direct fusion method (IDF) fuses monocular depth and binocular disparity through direct concatenation and convolution at each iteration. Compared to this approach, our iterative local fusion (ILF) is more robust to noise in binocular disparity, resulting in superior performance. The post-fusion method (PF) fuses monocular depth with the optimized binocular disparity from the previous iteration without registration. Compared to this approach, our global fusion (GF) achieves better compatibility between monocular depth and binocular disparity, mitigating the noise caused by scale ambiguity during fusion. 
Our iterative local fusion and global fusion modules further enhance performance and improve model robustness with the monocular encoder. It is also noted that the Baseline time cost is 0.32s, while our model's is 0.4s. Though it involves a VTF model, thanks to the elegant and controllable design, our model barely raises the time cost.
% Our iterative local fusion and global fusion modules further enhance performance and improve model robustness when combined with the monocular encoder. It is also noted that the time cost of the Baseline is 0.32s while our model is 0.4s. Even though it involves a VTF model, benefiting from the elegant and controllable design, our model barely raises the time cost.
% It is worth noting that the re-trained Baseline (RAFT-Stereo) shows a 1-point decrease compared to the official results in Table \ref{tab: mixed datasets}. This discrepancy is likely attributable to the use of A40 GPUs for training, which may lead to a minor performance drop (1-2 points) compared to other GPUs. Due to the unavailability of other GPUs, we opted to use 4 A40 GPUs to reduce convergence time, whose results are enough to effectively demonstrate the validity of our method.

\noindent\textbf{The Analysis of Iterative Local Fusion.}
We also analyze the specific configurations of iterative local fusion. As shown in Table \ref{tab: ILF}, the fixed weights in the LBP-like operation have a slight impact on performance, with a kernel size of ${3,5}$ providing optimal results. We also try to use convolutions with learnable weights to replace LBP-like convolutions. Comparing L(4), L(9-10) with L(6-7), we find that fixed-weight convolutions are more robust than learnable convolutions, and deeper learnable convolutions produce worse results. This is because limited data makes monocular-related learning unreliable for generalization, whereas manually designed convolutions incorporate prior knowledge and are less affected by data bias. Using a sigmoid function after LBP-like convolutions further improves overall performance. The amplitude parameter does not show a significant influence.
We also visualize the local ordering map in Figure \ref{Fig: vis lbp}. The local ordering maps of predicted disparity gradually become similar to the result of monocular depth as the iteration increases. For more visualizations, please refer to our supplemental materials.

\noindent\textbf{The Analysis of Components in Global Fusion.}
We analyze the specific configurations of global fusion, as shown in Table \ref{tab: GF}. Comparing G(1) with G(2), global fusion achieves nearly a 1-point improvement in the Bad 2.0 metric after registration. Comparing G(2) with G(3), learning confidence with more information enhances overall performance. Comparing MonoDepth and G(3), the fused results are more robust to monocular depth. We also visualize the registration parameters $\{ a, b \}$ in Figure \ref{Fig: vis reg}. $\{ a, b \}$ are changed in different areas but remain inconsistent for every pixel.  Due to page limitations, please refer to our supplementary materials for additional failure case analysis and future work discussion.

% \subsection{Limitations and Discussion}
% While our method has demonstrated significant improvements, there are still many challenges in the wild world, such as completely transparent glass, overly close mirrors, and distant objects. Furthermore, it is also worthy of future exploration to build a simulator for stereo data generation using pre-trained large models. Due to page limitations, please refer to our supplementary materials for additional failure case analysis and future work discussion.

%% file: sec/5_conclusion.tex
\section{Conclusion}
% \label{sec:conclusion}
In this paper, we dived into the fusion of monocular priors from VTF stereo matching and found three main problems limiting the fusion process. We proposed a binary local ordering map to unify the relative monocular depth and absolute disparity map. It also guided the fusion between monocular and binocular depth information in an explicit and controllable manner. Besides, we formulated the optimization of the disparity map as a registration process to monocular depth, which can adaptively and globally align the two kinds of depth maps. We designed a network to extract the unbiased monocular priors from the VFM and and leveraged the above two modules to fully exploit the unbiased monocular prior to the stereo matching pipeline to improve generalization in the ill-posed regions. Benefiting from the explicit design, our method barely increased the computation cost. Experimental results demonstrated the effectiveness of our method, with a significant improvement of 10 points on Booster and an error reduction of more than half on Middlebury and ETH3D, without using additional stereo data or data augmentation.

\textbf{Acknowledgement} This work was supported by the Shenzhen Science and Technology Program under Grant No. JCYJ20241202130548062, the Natural Science Foundation of Shenzhen under Grant No. JCYJ20230807142703006, the Natural Science Foundation of China (NSFC) under Grants No. 62176021 and No. 6217204, and the Key Research Platforms and Projects of the Guangdong Provincial Department of Education under Grant No.2023ZDZX1034.
% \vspace{-0.4cm}
\clearpage

%% file: sec/6_supp.tex
\clearpage
\setcounter{page}{1}
\maketitlesupplementary
% \makeauthorsupplementary

% % \vspace{-11mm}
% \begin{figure*}[htbp]
%   \centering
%   \href{https://huggingface.co/spaces/AdamYao/Diving-into-the-Fusion-of-Monocular-Priors-for-Generalized-Stereo-Matching}{%
%     \includegraphics[height=12pt]{./Figure/hf_demo}
%   }%
%   \hspace{0.8em}%
%   \href{https://github.com/YaoChengTang/Diving-into-the-Fusion-of-Monocular-Priors-for-Generalized-Stereo-Matching}{%
%     \includegraphics[height=12pt]{./Figure/github_repo}
%   }%
%   \hspace{0.8em}%
%   \href{https://drive.google.com/drive/folders/1PaQPOzzDajlnFfNm2fBKZsfJKb2XPejd?usp=sharing}{%
%     \includegraphics[height=12pt]{./Figure/model_weights}
%   }
% \end{figure*}
% % \vspace{-1em}

% \begin{figure*}
%   % \vspace{-1mm}
%   \centering
%   \includegraphics[width=\textwidth]{Figure/Flicker.jpg}
%   \vspace{-6mm}
%   \captionof{figure}{The visualization of results on the Flicker1024 dataset.}
%   \label{fig:flicker}
% \end{figure*}

% % \tableofcontents  %在指定位置生成目录
% {
%  \small
%  \tableofcontents
% }

\begin{figure}[htbp]
\centering
    \includegraphics[width=.48\textwidth]{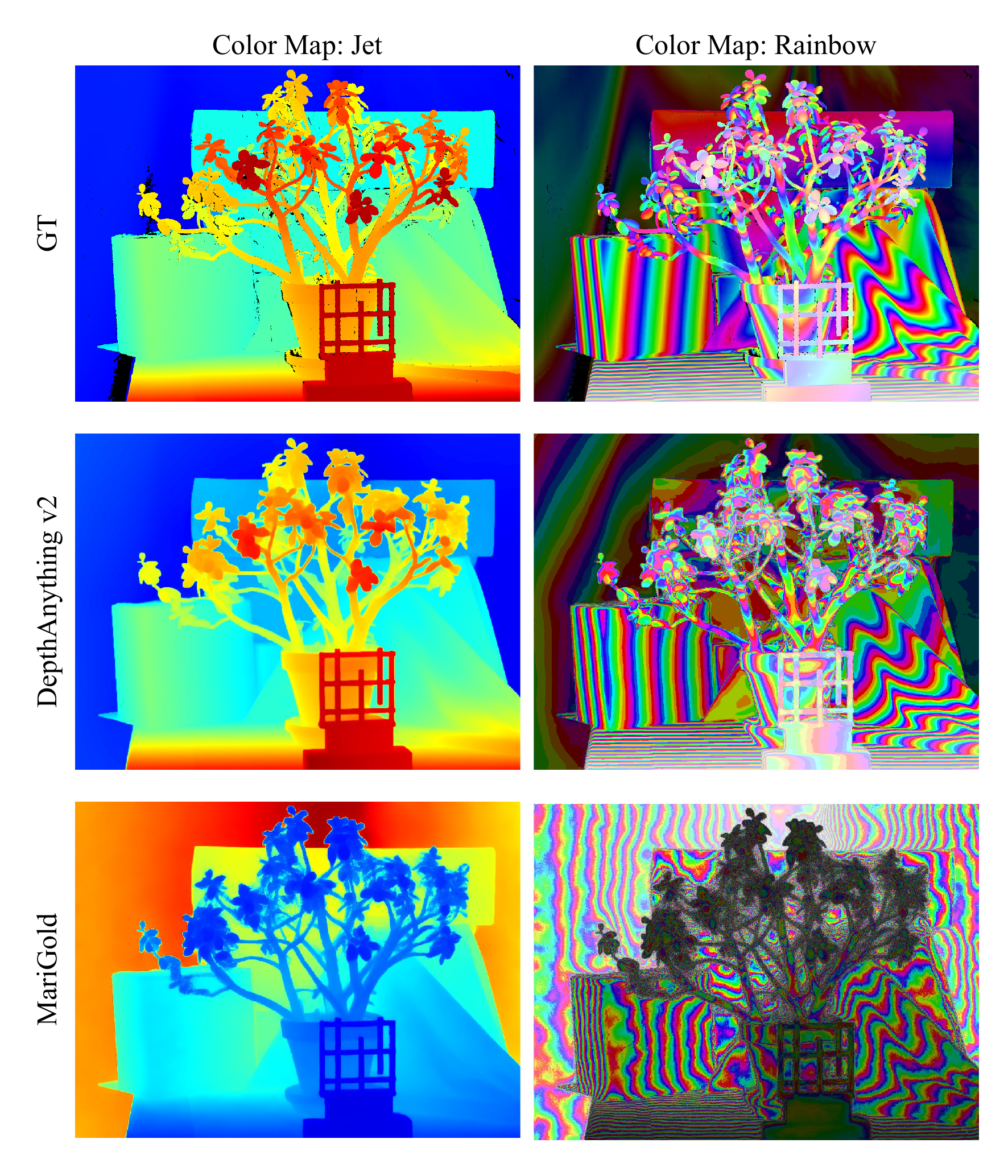}
    \caption{The visualization of results. We use two kinds of colormap to visualize the disparity map.}
    % \vspace{-0.3cm}
    \label{Fig: vis-mono-depth}
\end{figure}

\noindent The training and testing codes for all experiments, including the ablation study, are available in our project. For reproducibility, we strongly recommend referring to our project.

\section{Visualization On Flicker1024}
We present visualization results demonstrating the generalization capability of our model from the synthetic SceneFlow dataset to the real-world Flickr1024 dataset \cite{Flickr1024}. As shown in Figure~\ref{fig:flicker}, our model performs robustly across diverse scenarios, including large outdoor and indoor scenes, thin and small objects, strong lighting interference and low-light conditions, as well as challenging materials such as glass windows, walls, and bottles.
% We present the visualization results of our model generalized from the SceneFlow dataset to the Flciker1024 dataset \cite{Flickr1024}. As shown in Figure \ref{fig:flicker}, we show the effectiveness of our model in various scenarios, from large outdoor to indoor scenes, from thin objects to small objects, from strong light interference to dark scenes, from common materials to glass windows, walls, and bottles.

\section{Intuition behind Monocular Depth Model}
We choose DepthAnything v2 \cite{yang2024depthv2} over Marigold \cite{ke2024repurposing} because of the superior continuity of its depth maps. As shown in Figure \ref{Fig: vis-mono-depth}, DepthAnything v2 provides depth maps with better continuity than Marigold, especially in fine-grained regions. The depth maps from Marigold contain considerable noise, while those from DepthAnything v2 are much cleaner.

\begin{table*}[htbp]
  \centering
  \renewcommand\arraystretch{1.1}
  \setlength{\tabcolsep}{.8mm}
    \begin{tabular}{c|c|cccc|cccc}
    \hline
    \multirow{3}[3]{*}{Method} & \multirow{3}[3]{*}{ \tabincell{c}{Additional \\ Data/Aug} } & \multicolumn{8}{c}{Booster} \\
        \cline{3-10}          &       & \multicolumn{4}{c|}{Class 0} & \multicolumn{4}{c}{Class 1} \\
                  &       & \multicolumn{1}{c}{EPE} & \multicolumn{1}{c}{bad 2.0} & \multicolumn{1}{c}{bad 3.0} & \multicolumn{1}{c|}{bad 5.0} & \multicolumn{1}{c}{EPE} & \multicolumn{1}{c}{bad 2.0} & \multicolumn{1}{c}{bad 3.0} & \multicolumn{1}{c}{bad 5.0} \\
    \hline
    Mocha-Stereo 192\cite{chen2024mocha} &       & 1.30  & 6.93 & 5.54 & 4.18 & 2.91  & 23.05 & 17.67 & 13.45 \\
    Mocha-Stereo 320\cite{chen2024mocha} &    &1.20    &6.18    &4.84    &3.53    &2.88    &22.83    &17.34    &12.98\\
    ELFNet \cite{lou2023elfnet} &           &2.97    &14.08    &11.38    &8.80    &5.67    &24.68    &19.00    &14.42   \\
    Selective-RAFT \cite{wang2024selective} &    &1.35 &8.06 &6.01 &4.01 &3.37 &27.37 &21.87 &17.19 \\
    Selective-IGEV 192\cite{wang2024selective} &    &1.46    &8.03    &6.19    &4.66    &3.61    &25.57    &20.05    &15.93    \\
    Selective-IGEV 320\cite{wang2024selective} &        &1.31    &7.27    &5.39    &3.81    &3.51    &25.05    &19.39    &15.18 \\
    IGEV 192\cite{xu2023iterative} &           &1.17    &6.67    &4.84    &3.46    &3.76    &25.46    &20.26    &16.39    \\
    IGEV 320\cite{xu2023iterative} &    &1.00    &6.07    &4.37    &2.82    &3.60    &24.69    &19.46    &15.70\\
    NMRF \cite{guan2024neural} &           &2.76    &17.43    &13.21    &9.51    &4.60    &32.81    &26.08    &19.84    \\
    NerfStereo \cite{tosi2023nerf} & \checkmark     &\textbf{0.73}    &\textbf{4.07}    & \textbf{2.55}    &\textbf{1.47}    &2.41    &18.67    &13.92    &10.56 \\
    RAFTstereo \cite{lipson2021raft} &           &1.14    &5.84    &4.39    &3.08    &3.66    &25.34    &19.35    &14.37     \\
    \hline
    RAFT-Stereo + ME &        &0.96 &6.57 & 5.24 & 3.93 & 1.81 &13.68 &8.77 & 5.98 \\
    \textbf{Ours} &           &0.79    &5.90    &4.57    &3.17    &\textbf{1.53}    &\textbf{12.67}    &\textbf{7.80}    &\textbf{4.88} \\
    \hline
    \end{tabular}%
    % \captionsetup{skip=3pt} % 设置特定表格的标题与正文距离
  % \vspace{-3mm}
  \caption{Generalization from SceneFlow dataset to Booster dataset in quarter resolution and balanced set. ME represents our monocular encoder module. All results are evaluated in the same metrics and settings. The 192 and 320 represent the maximum disparity range used in each model.}
  % \vspace{-1mm}
  \label{tab: booster_0_1}%
\end{table*}%

\begin{table*}[htbp]
  \centering
  \renewcommand\arraystretch{1.1}
  \setlength{\tabcolsep}{.8mm}
    \begin{tabular}{c|c|cccc|cccc}
    \hline
    \multirow{3}[3]{*}{Method} & \multirow{3}[3]{*}{ \tabincell{c}{Additional \\ Data/Aug} } & \multicolumn{8}{c}{Booster} \\
        \cline{3-10}          &       & \multicolumn{4}{c|}{Class 2} & \multicolumn{4}{c}{Class 3} \\
                  &       & \multicolumn{1}{c}{EPE} & \multicolumn{1}{c}{bad 2.0} & \multicolumn{1}{c}{bad 3.0} & \multicolumn{1}{c|}{bad 5.0} & \multicolumn{1}{c}{EPE} & \multicolumn{1}{c}{bad 2.0} & \multicolumn{1}{c}{bad 3.0} & \multicolumn{1}{c}{bad 5.0} \\
    \hline
    Mocha-Stereo 192\cite{chen2024mocha} &       & 15.68  & 53.56 & 46.23 & 37.77 & 9.45  & 66.44 & 57.96 & 45.73 \\
    Mocha-Stereo 320\cite{chen2024mocha} &       &15.05    &53.88    &46.63    &37.62    &9.21    &65.88    &57.30 &44.65\\
    ELFNet \cite{lou2023elfnet} &          &22.74    &78.89    &74.81    &69.70  &9.03  &72.07  &62.73  &49.82   \\
    Selective-RAFT \cite{wang2024selective} &     &16.12 &55.66 &49.87 &43.04  & 10.34 & 69.84 & 61.64 & 49.55\\
    Selective-IGEV 192\cite{wang2024selective} &     &20.41    &57.55    &49.78    &42.86 & 9.50   & 66.85 & 58.9  & 47.15\\
    Selective-IGEV 320\cite{wang2024selective} &       &19.81    &57.35    &49.27    &42.10    &9.29    &66.02    &57.91    &45.86\\
    IGEV 192\cite{xu2023iterative} &          &18.55    &54.64    &46.45    &37.79 & 10.00 & 68.96 & 61.14 & 49.51\\
    IGEV 320\cite{xu2023iterative} &       &18.00    &54.50    &46.05    &37.72    &9.74    &68.55    &60.49    &48.22\\
    NMRF \cite{guan2024neural} &           &17.36    &56.34    &48.33    &38.18 & 10.36 &  70.92 & 60.93 & 47.16\\
    NerfStereo \cite{tosi2023nerf} & \checkmark      &17.92    &45.67    &40.39    &35.19 & 8.88 & 62.67 & 53.35 & 41.79  \\
    RAFTstereo \cite{lipson2021raft} &          &18.58    &54.00    &47.52    &40.44 & 9.79 & 67.69 & 59.31 & 47.40 \\
    \hline
    RAFT-Stereo + ME &       &\textbf{5.16} & 24.38 & 19.01 & 14.58 & 8.97 & 64.84 & 56.05 & 43.95\\
    \textbf{Ours} &           &5.32    &\textbf{23.34}    &\textbf{17.62}    &\textbf{13.50} & \textbf{7.93} & \textbf{59.83} & \textbf{50.36} & \textbf{38.44} \\
    \hline
    \end{tabular}%
    % \captionsetup{skip=3pt} % 设置特定表格的标题与正文距离
  \caption{Generalization from SceneFlow dataset to Booster dataset in quarter resolution and balanced set. ME represents our monocular encoder module. All results are evaluated in the same metrics and settings. The 192 and 320 represent the maximum disparity range used in each model.}
  \vspace{-1mm}
  \label{tab: booster_2_3}%
\end{table*}%

\section{More Results on Booster}
\begin{figure*}[htbp]
\centering
    \includegraphics[width=0.8\textwidth]{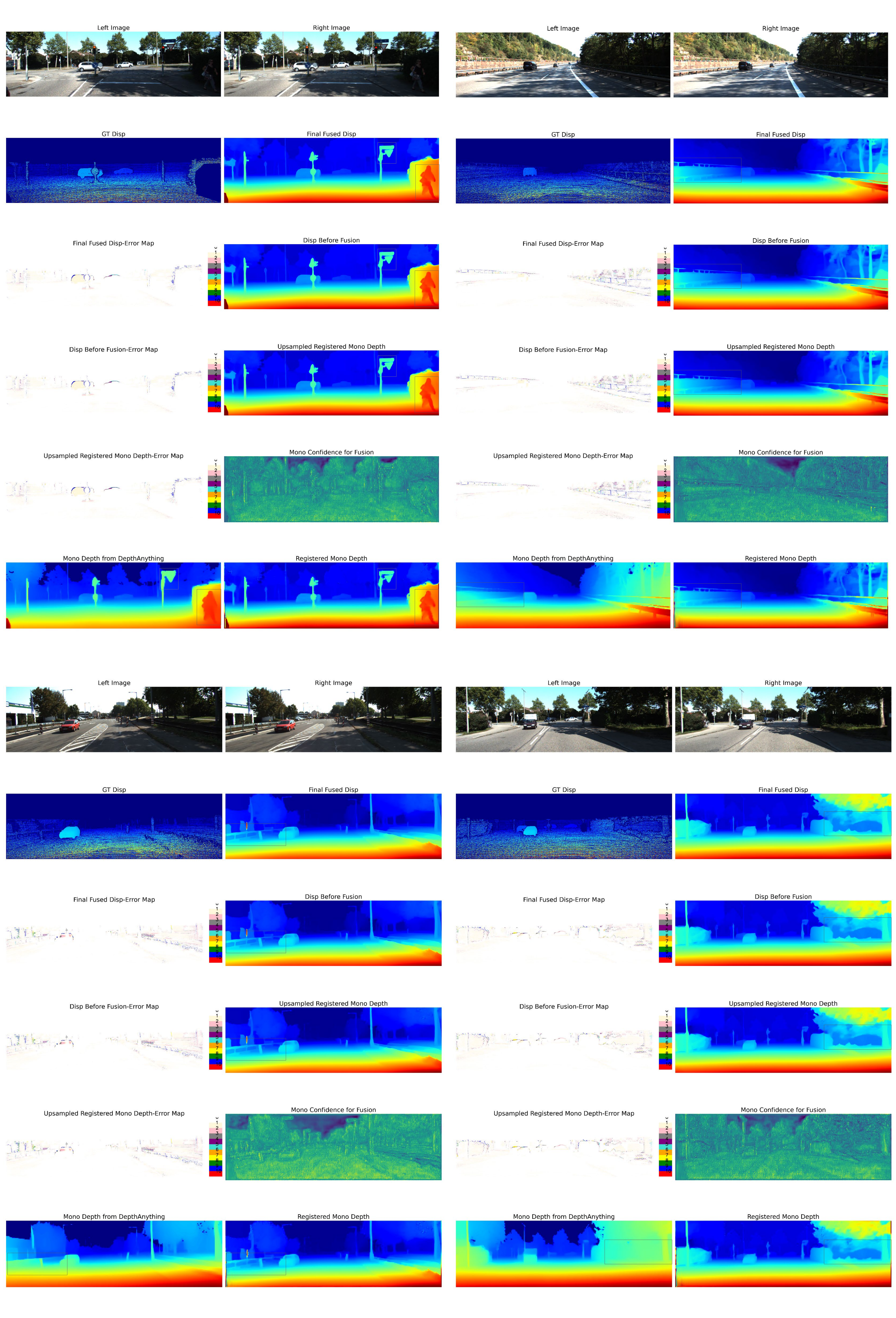}
    \caption{The visualization of binocular disparity and monocular depth. The regions highlighted with gray boxes demonstrate that stereo matching excels at capturing fine-grained details, whereas monocular depth estimation performs better in perceiving overall shapes. The mono depth from DepthAnything is scale ambiguity but not absolute depth before registration.}
    \label{Fig: visup1}
\end{figure*}

We provide additional results on the Booster dataset across various material types. From class $0$ to $3$, the materials become increasingly transparent and/or specular. As shown in Tables \ref{tab: booster_0_1} and \ref{tab: booster_2_3}, our method outperforms state-of-the-art approaches on transparent and/or specular objects (classes $1$ to $3$), while achieving comparable results in normal regions (class $0$). The normal regions of the Booster dataset mainly consist of regular objects, flat surfaces, or highly textured areas. Consequently, NerfStereo, which incorporates additional stereo data, performs particularly well in these regions. This indicates that stereo matching effectively captures fine-grained details, whereas monocular depth estimation excels in perceiving coarse shapes. As illustrated in Figures \ref{Fig: visup1} and \ref{Fig: visup2}, binocular disparity provides greater detail compared to monocular depth. Our method disentangles monocular depth and binocular disparity, allowing the model to leverage both monocular and stereo data, and explore the fusion of monocular priors effectively.

\section{Additional Training Data}
We evaluate the scalability of our model by incorporating additional training data from the TranScene dataset \cite{Liu2025MultiLabelSM}, a synthetic dataset specifically designed for multi-label transparent scenes. In our experiments, we use labels with the largest disparity in transparent regions. It should be noted that, this time, our model is trained end-to-end using weights pretrained on the SceneFlow dataset, without using any multi-stage training strategy. As shown in Tables \ref{tab:add_data_main} and \ref{tab:add_data_sub}, incorporating the additional data leads to consistent performance improvements across all evaluated metrics, with particularly notable gains in transparent regions. Furthermore, our model's performance on common scenes (e.g., non-transparent regions) not only remains stable but also shows slight improvement. These results highlight the scalability potential of our model when augmented with additional large data.

\begin{table*}[htbp]
\centering
\renewcommand\arraystretch{1.}
\begin{tabular}{c|cc|cc|cc}
\hline
\multirow{2}{*}{Metric} 
& \multicolumn{2}{c|}{ALL} 
& \multicolumn{2}{c|}{Trans} 
& \multicolumn{2}{c}{NoTrans} \\
& Ours & Ours+TranScene & Ours & Ours+TranScene & Ours & Ours+TranScene \\
\hline
EPE   & 2.26 & \textbf{1.24} & 7.93 & \textbf{5.67} & 1.52 & \textbf{0.75} \\
RMSE  & 5.60 & \textbf{4.19} & 11.03 & \textbf{8.42} & 3.93 & \textbf{3.07} \\
2px   & 11.02 & \textbf{7.91} & 59.83 & \textbf{46.78} & 6.98 & \textbf{4.77} \\
3px   & 8.59 & \textbf{5.97} & 50.36 & \textbf{38.55} & 4.97 & \textbf{3.23} \\
5px   & 6.60 & \textbf{4.52} & 38.44 & \textbf{28.65} & 3.64 & \textbf{2.29} \\
6px   & 6.00 & \textbf{4.08} & 33.87 & \textbf{25.41} & 3.27 & \textbf{2.01} \\
8px   & 5.35 & \textbf{3.44} & 27.56 & \textbf{21.30} & 2.89 & \textbf{1.59} \\
\hline
\end{tabular}
% \vspace{-3mm}
\caption{Generalization from the SceneFlow dataset to the Booster dataset in quarter resolution and balanced set. `All', `Trans', and `NonTrans' represent all regions, transparent regions, and nontransparent regions, respectively.}
\label{tab:add_data_main}
% \vspace{-3mm}
\end{table*}

\begin{table*}[htbp]
\centering
\renewcommand\arraystretch{1.}
\begin{tabular}{c|cc|cc|cc|cc}
\hline
\multirow{2}{*}{Metric} 
& \multicolumn{2}{c|}{Class 0} 
& \multicolumn{2}{c|}{Class 1} 
& \multicolumn{2}{c|}{Class 2} 
& \multicolumn{2}{c}{Class 3} \\
& Ours & Ours+TranScene & Ours & Ours+TranScene & Ours & Ours+TranScene & Ours & Ours+TranScene \\
\hline
EPE   & 0.79 & \textbf{0.75} & 1.53 & \textbf{1.40} & 5.32 & \textbf{1.62} & 7.93 & \textbf{5.67} \\
RMSE  & 3.02 & \textbf{2.99} & 4.70 & \textbf{4.74} & 6.39 & \textbf{2.26} & 11.03 & \textbf{8.42} \\
2px   & 5.90 & \textbf{5.15} & 12.67 & \textbf{9.17} & 23.34 & \textbf{13.51} & 59.83 & \textbf{46.78} \\
3px   & 4.57 & \textbf{4.08} & 7.80 & \textbf{5.63} & 17.62 & \textbf{10.23} & 50.36 & \textbf{38.55} \\
5px   & 3.17 & \textbf{3.00} & 4.88 & \textbf{3.80} & 13.50 & \textbf{7.40} & 38.44 & \textbf{28.65} \\
6px   & 2.58 & \textbf{2.59} & 3.96 & \textbf{3.37} & 12.80 & \textbf{6.50} & 33.87 & \textbf{25.41} \\
8px   & 1.45 & \textbf{1.73} & 3.14 & \textbf{2.86} & 12.15 & \textbf{4.93} & 27.56 & \textbf{21.30} \\
\hline
\end{tabular}
% \vspace{-3mm}
\caption{Generalization from the SceneFlow dataset to the Booster dataset in various regions.}
\label{tab:add_data_sub}
% \vspace{-3mm}
\end{table*}

% Table generated by Excel2LaTeX from sheet 'Sheet2'
\begin{table*}[h]
  \centering
  % \scalebox{0.8}{
  \renewcommand\arraystretch{1.}
  \setlength{\tabcolsep}{1.3mm}
    \begin{tabular}{c|ccccc}
    \hline
          & 750$\times$2484 & 1125$\times$3726 & 1500$\times$4968 & 1688$\times$5589 & 1875$\times$6210 \\
    \hline
    RAFTStereo reg \citep{lipson2021raft} & 2268.35 & 6023.82 & 10795.02 & 14299.5 & 19666.78 \\
    RAFTStereo alt \citep{lipson2021raft} & 1715.8 & 4151.8 & 6466.7 & 8157.96 & 11177.66 \\
    IGEV 384 \citep{xu2023iterative} & 2816.46 & 7290.82 & 14484.61 & 18810.14 & - \\
    IGEV 640 \citep{xu2023iterative} & 3167.46 & 8475.43 & 17366.83 & -     & - \\
    Selective IGEV 384 \cite{wang2024selective} & 2960.34 & 7608.44 & 15035.55 & 19505.5 & - \\
    Selective IGEV 640 \cite{wang2024selective} & 3311.84 & 8793.07 & 18701.57 & -     & - \\
    Mocha-Stereo 384 \cite{chen2024mocha} & 5525.56 & 12986.73 & 24665.95 & -     & - \\
    Mocha-Stereo 640 \cite{chen2024mocha} & 6136.18 & 15056.66 & 29476.45 & -     & - \\
    \hline
    ours reg & 5031.07 & 8609.63 & 14088.98 & 17782.53 & 22279.12 \\
    ours alt & 3452.42 & 6745.23 & 9761.22 & 11641.82 & 13790.82 \\
    \hline
    \end{tabular}%
  % }
%   \vspace{-3mm}
  \caption{Memory comparison across different resolutions. We evaluate the memory consumption of each model, excluding the feature encoder module, to ensure a fair comparison of backbones during inference. reg: pre-computation of the entire cost volume, allowing for look-up operations at each iteration, alt: dynamically computing a thin cost volume at each iteration. 384/640: the maximum disparity range used for the resolution of 750$\times$2484. '-':  out of memory in our GPU.}
  \label{tab: memory}%
%   \vspace{-3mm}
\end{table*}%

\begin{table}[htbp]
  \centering
  \scalebox{0.95}{
  \renewcommand\arraystretch{1.}
  \setlength{\tabcolsep}{1.8mm}
    \begin{tabular}{c|c|c}
    \hline
    \multirow{2}[1]{*}{Exp} & \multicolumn{2}{c}{Middlebury (H)} \\
        \cline{2-3}          & epe   & bad 2.0 \\
    \hline
    Baseline + FE-DepthAnything & 3.26$\pm$0.03 & 28.73$\pm$0.28 \\
    Baseline + FE-MASt3R & 4.41$\pm$0.40 & 26.83$\pm$0.57 \\
    Baseline + ME + ILF + GF & 1.15$\pm$0.01 & 8.35$\pm$0.04 \\
    \hline
    \end{tabular}%
    }
  % \captionsetup{skip=3pt} % 设置特定表格的标题与正文距离
%   \vspace{-3mm}
  \caption{The effectiveness of each module. Baseline: RAFTStereo, ME: our monocular encoder, ILF: iterative local fusion, GF: our global fusion. FE-DepthAnything: replacing the original feature extractor with DepthANything v2. FE-MASt3R: replacing the original feature extractor with MASt3R.}
  \label{tab: backbone}%
  % \vspace{-0.8cm}
  \vspace{-3mm}
\end{table}%

\section{More Analysis about Memory}
We also compare our model to state-of-the-art methods in terms of memory consumption across different resolutions. To ensure a fair comparison of backbones during inference, we exclude the feature encoder module when evaluating each model's memory consumption. Notably, the memory consumption of IGEV becomes extremely high on the A40 GPU as the maximum disparity range increases. We suspect this may be a bug; therefore, we used a borrowed 4090 GPU for evaluations under the first four resolutions, while the evaluation under the last resolution was conducted on the A40 GPU.

As shown in Table \ref{tab: memory}, our method, along with RAFTStereo \citep{lipson2021raft}, maintains a slower growth rate in memory consumption compared to IGEV \citep{xu2023iterative}, Selective IGEV \cite{wang2024selective}, and Mocha \cite{chen2024mocha}. Compared to RAFTStereo, our method exhibits a similar memory consumption increase across resolutions due to the resizing operation required by DepthAnything v2.

\section{More Visualization}
\begin{figure*}[htbp]
\centering
    \includegraphics[width=0.8\textwidth]{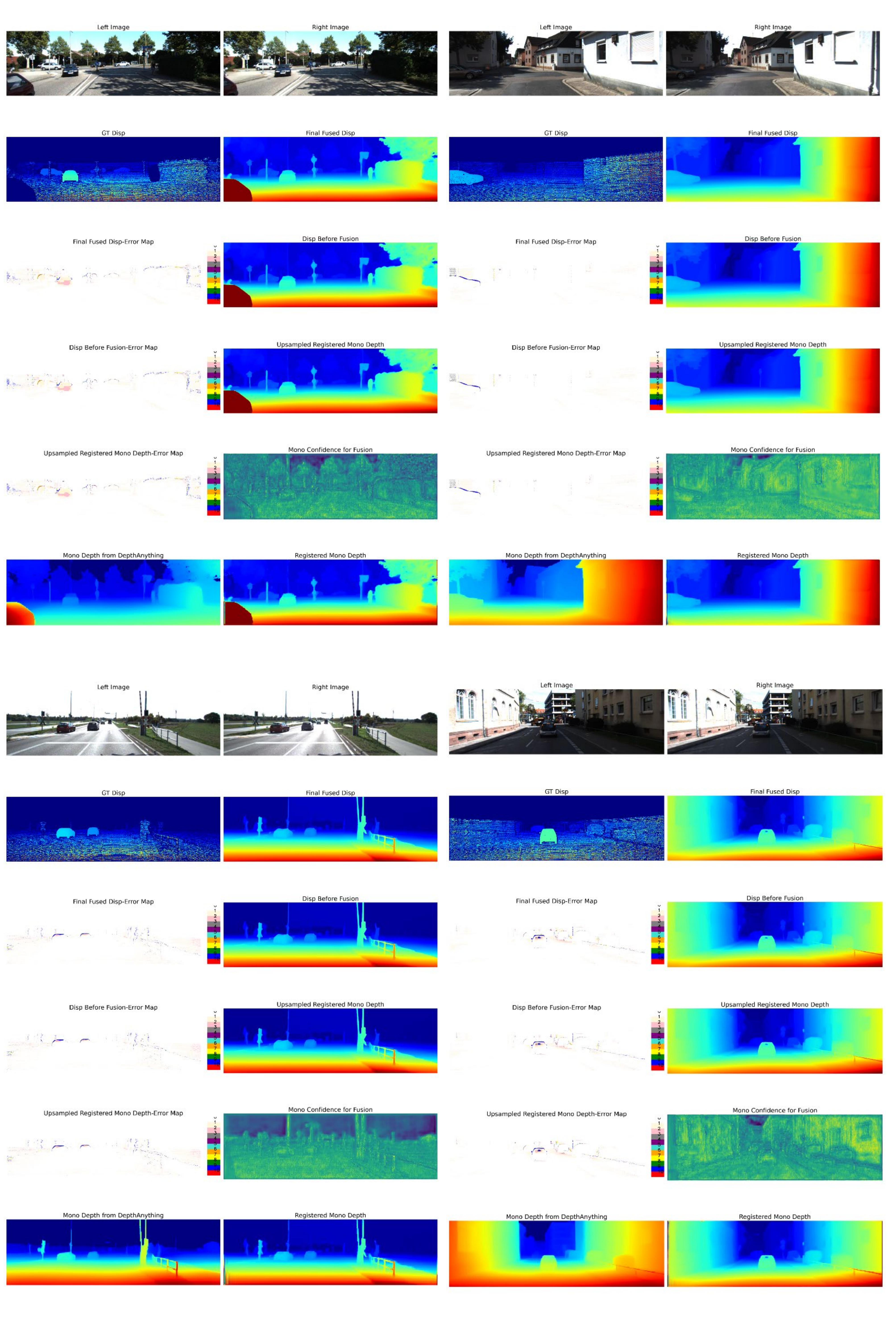}
    \caption{The visualization for generalized stereo matching.}
    \label{Fig: visup3}
\end{figure*}

\begin{figure*}[htbp]
\centering
    \includegraphics[width=0.8\textwidth]{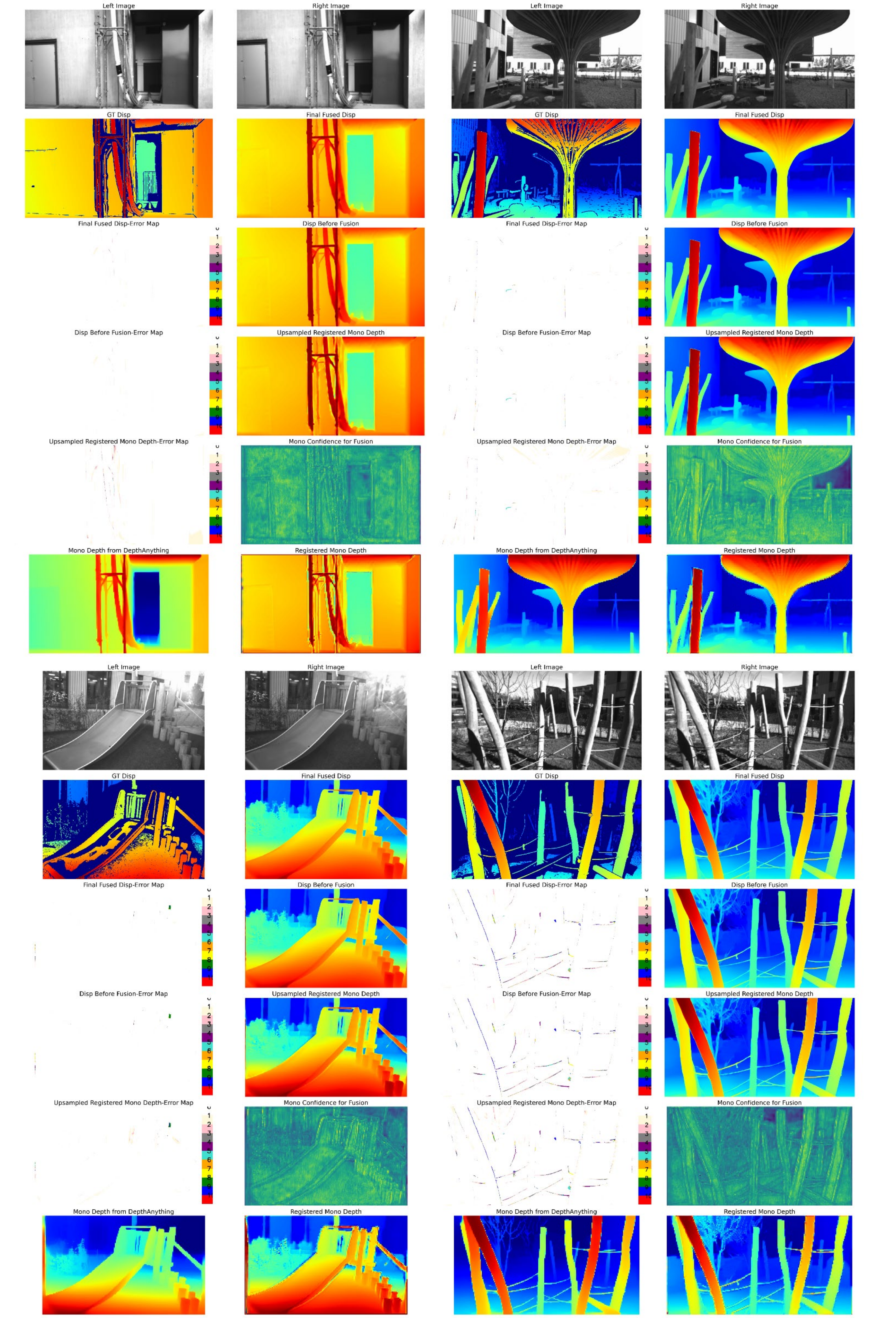}
    \caption{The visualization for generalized stereo matching.}
    \label{Fig: visup4}
\end{figure*}

\begin{figure*}[htbp]
\centering
    \includegraphics[width=0.8\textwidth]{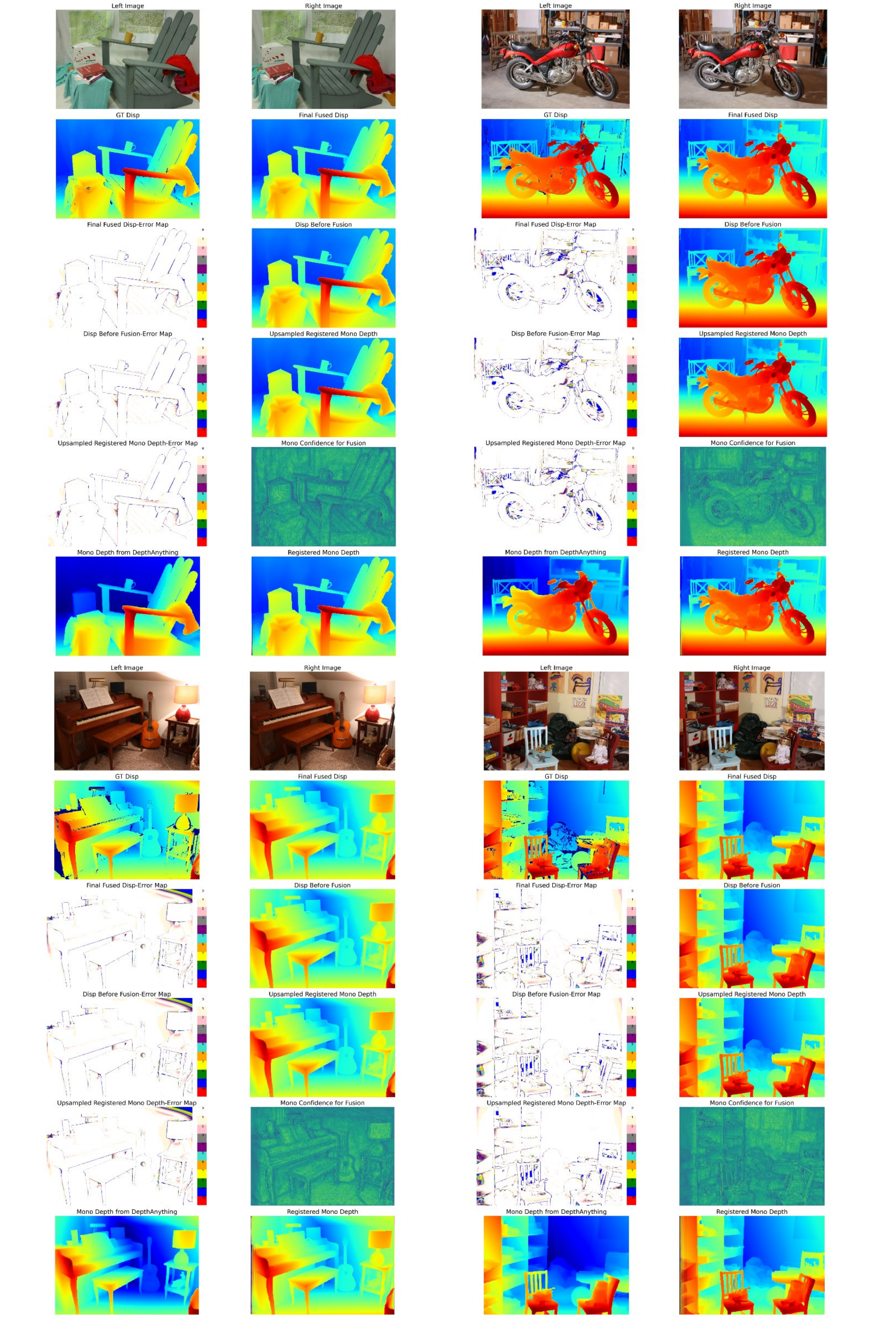}
    \caption{The visualization for generalized stereo matching.}
    \label{Fig: visup5}
\end{figure*}

\begin{figure*}[htbp]
\centering
    \includegraphics[width=0.8\textwidth]{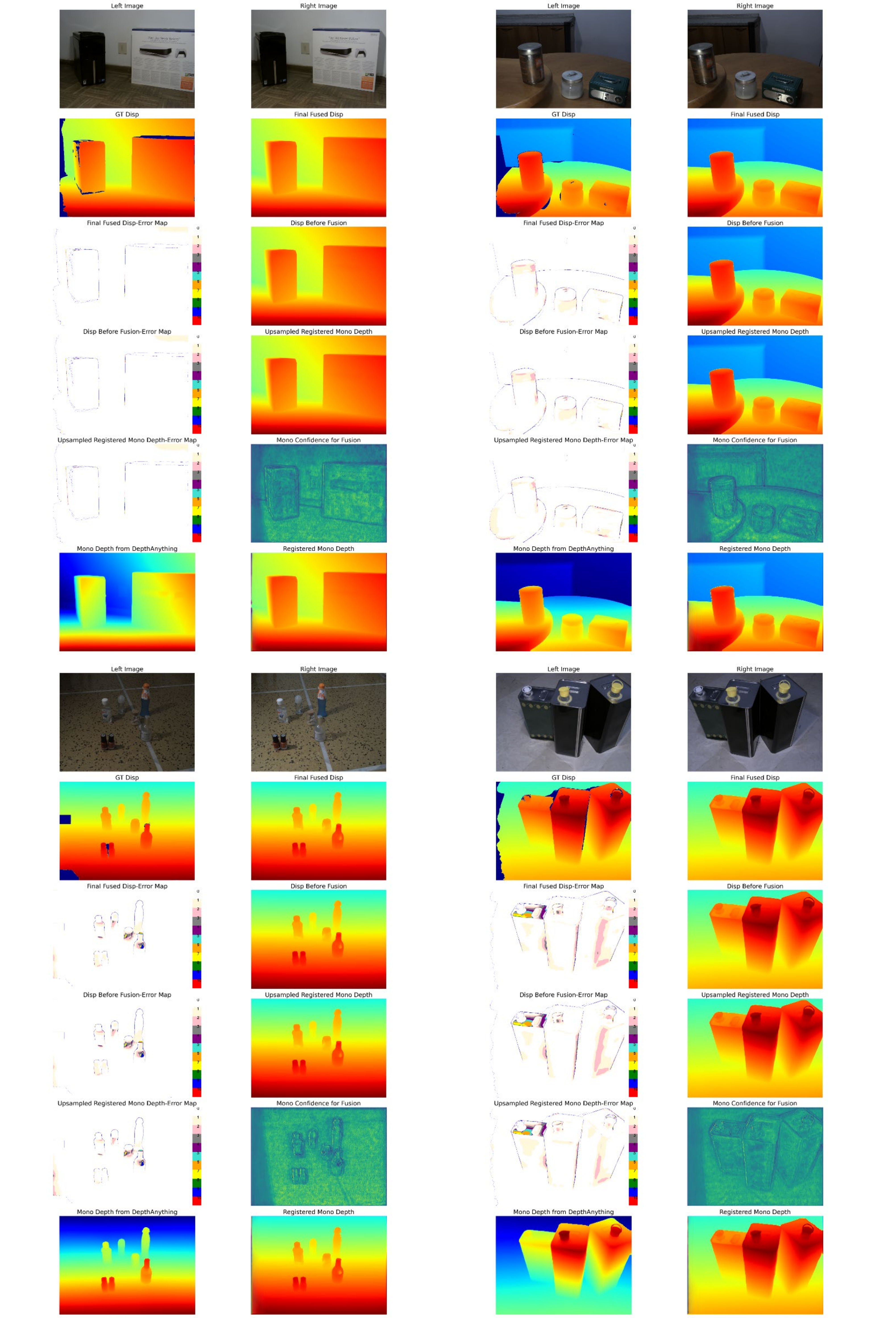}
    \caption{The visualization for generalized stereo matching.}
    \label{Fig: visup6}
\end{figure*}

\begin{figure*}[htbp]
\centering
    \includegraphics[width=0.7\textwidth]{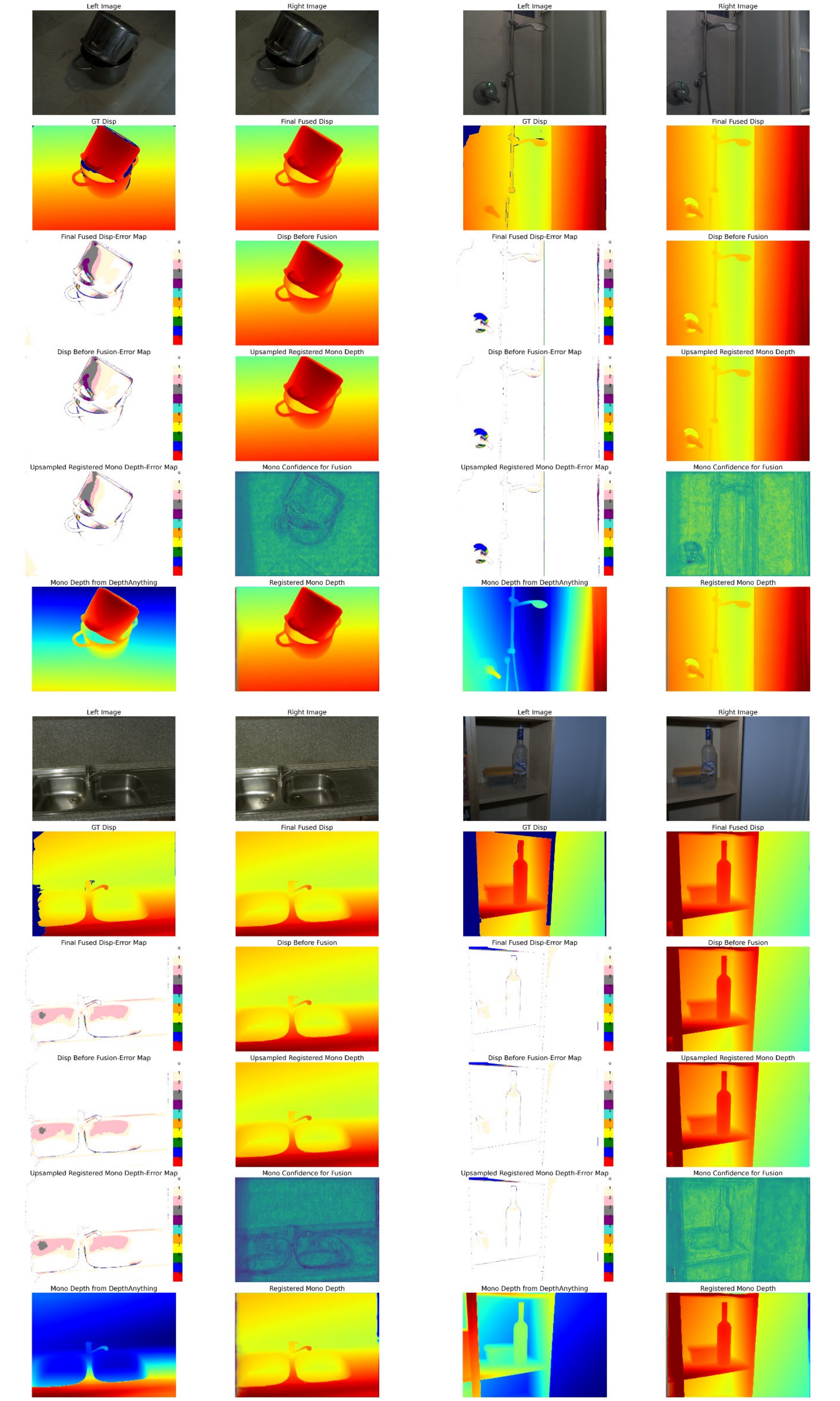}
    \caption{The visualization for generalized stereo matching.}
    \label{Fig: visup7}
\end{figure*}

We provide additional visualizations of generalized stereo matching in Figures \ref{Fig: visup3}, \ref{Fig: visup4}, \ref{Fig: visup5}, \ref{Fig: visup6}, and \ref{Fig: visup7}. The visualizations span a variety of environments, ranging from open outdoor scenes (e.g., driving scenarios), to semi-open outdoor scenes (e.g., playgrounds), and to enclosed indoor scenes (e.g., rooms, tables). The results demonstrate that our method generalizes effectively to the wild world, achieving strong performance even when trained only on a limited amount of synthetic stereo data.

\section{Ablation Study}

\subsection{More Analysis of Backbone}
In addition to replacing the context network with the pre-trained DepthAnything v2 \cite{yang2024depthv2}, we also experimented with replacing the feature extractor for cost volume construction using DepthAnything v2 \cite{yang2024depthv2} and MASt3R \cite{mast3r_arxiv24,dust3r_cvpr24}. As shown in Table \ref{tab: backbone}, the results become worse after replacing the feature extractor for cost volume construction with DepthAnything v2 or MASt3R. 
Moreover, a bug with the A40 GPU causes memory issues when converting the alternate correlation function from dot product to Euclidean distance during training. Therefore, the model with MASt3R was trained using the original correlation function with dot product, where additional learnable convolution layers are further used after MASt3R for feature extraction.
% Moreover, there is a bug with the A40 GPU that causes memory issues when converting the alternate correlation function from dot product to Euclidean distance during training. Therefore, the model with MASt3R was trained using the original correlation function with dot product, where additional learnable convolution layers are further used after MASt3R for feature extraction.

\subsection{More Analysis of Iterative Local Fusion}
We provide additional visualizations of the intermediate results from the iterative local fusion process in Figures \ref{Fig: vis-inter1}, \ref{Fig: vis-inter2}, \ref{Fig: vis-inter3}, \ref{Fig: vis-inter4}, \ref{Fig: vis-inter5}, \ref{Fig: vis-inter6}, \ref{Fig: vis-inter7}, and \ref{Fig: vis-inter8}. As the iterations progress, the ordering maps generated from binocular disparity gradually become smoother. The convolution layers learn the differences between ordering maps generated from binocular disparity and monocular depth, allowing the guidance to focus more effectively on non-smooth regions, thereby significantly affecting disparity update.

\subsection{More Analysis of Components in Global Fusion}
\begin{figure*}[htbp]
\centering
    \includegraphics[width=0.8\textwidth]{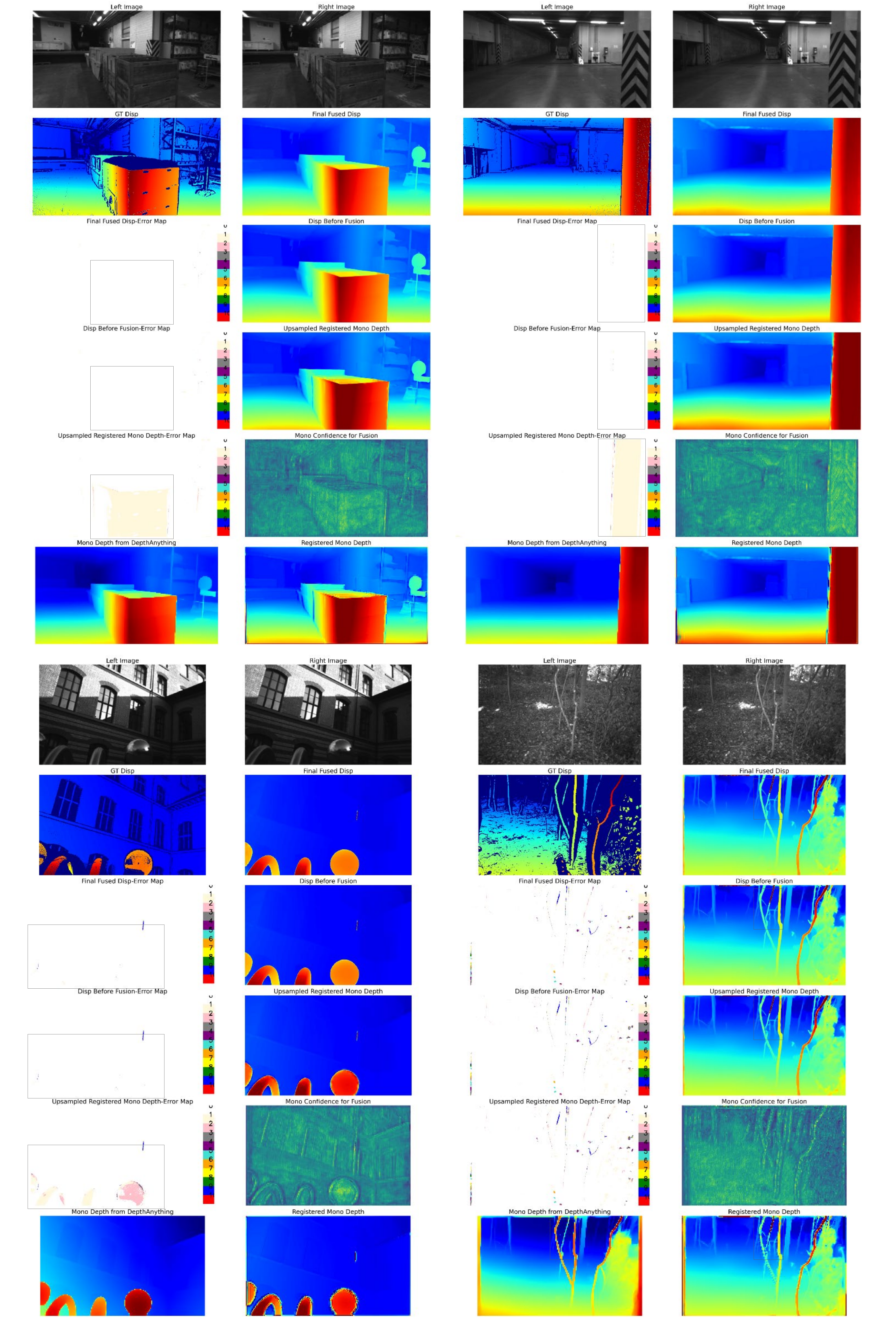}
    \caption{The visualization of binocular disparity and monocular depth. The regions highlighted with gray boxes demonstrate that stereo matching excels at capturing fine-grained details, whereas monocular depth estimation performs better in perceiving overall shapes. The mono depth from DepthAnything is scale ambiguity but not absolute depth before registration.}
    \label{Fig: visup2}
\end{figure*}

\begin{figure*}[htbp]
\centering
    \includegraphics[width=0.9\textwidth]{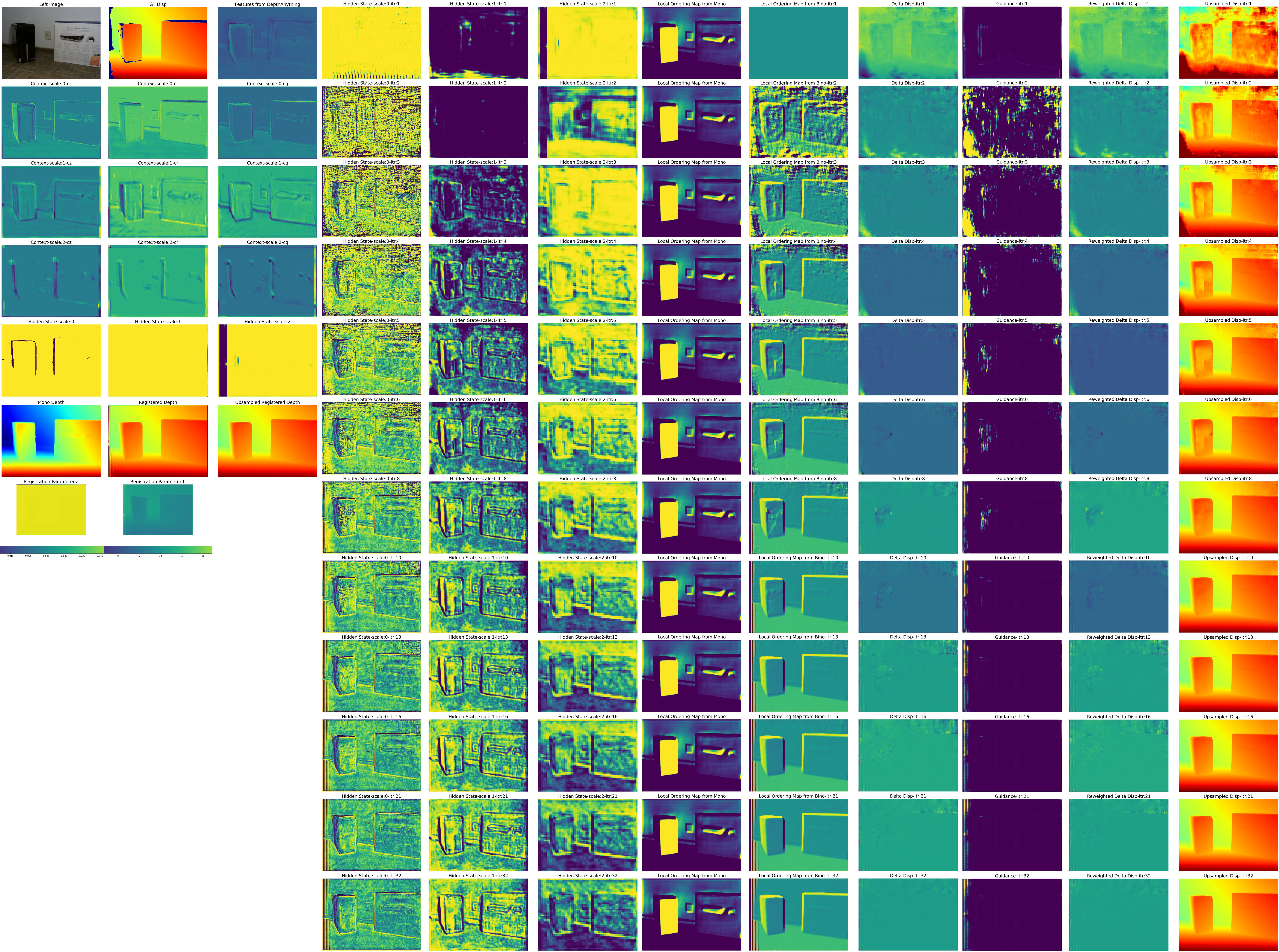}
    \caption{The visualization of intermediate results. $itr$: the current iteration. $cz, cr, cq$: context used in GRU. $scale$: scale $0\sim2$ represents resolution from high to low.}
    \label{Fig: vis-inter1}
\end{figure*}

\begin{figure*}[htbp]
\centering
    \includegraphics[width=0.9\textwidth]{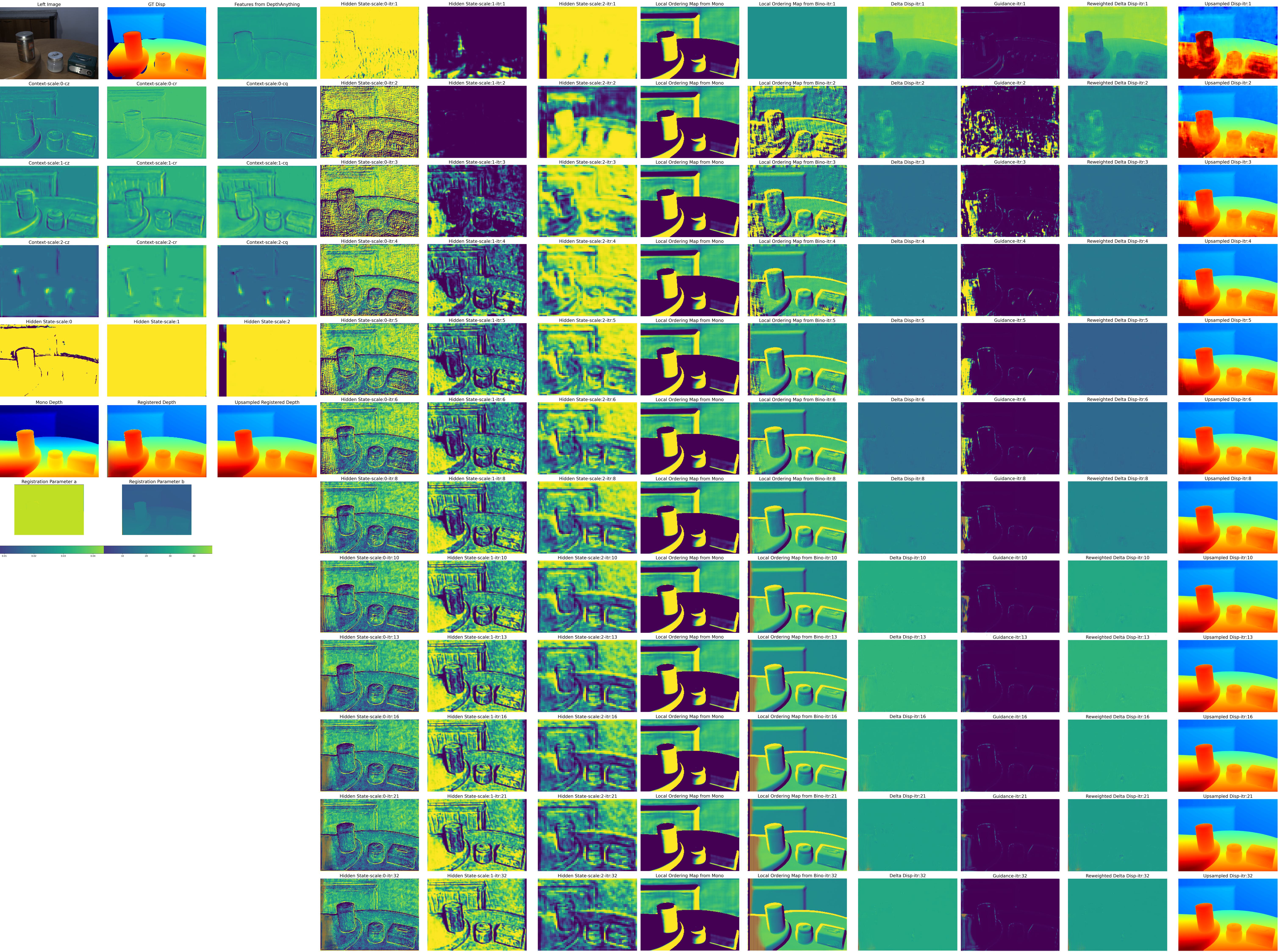}
    \caption{The visualization of intermediate results. $itr$: the current iteration. $cz, cr, cq$: context used in GRU. $scale$: scale $0\sim2$ represents resolution from high to low.}
    \label{Fig: vis-inter2}
\end{figure*}

\begin{figure*}[htbp]
\centering
    \includegraphics[width=0.9\textwidth]{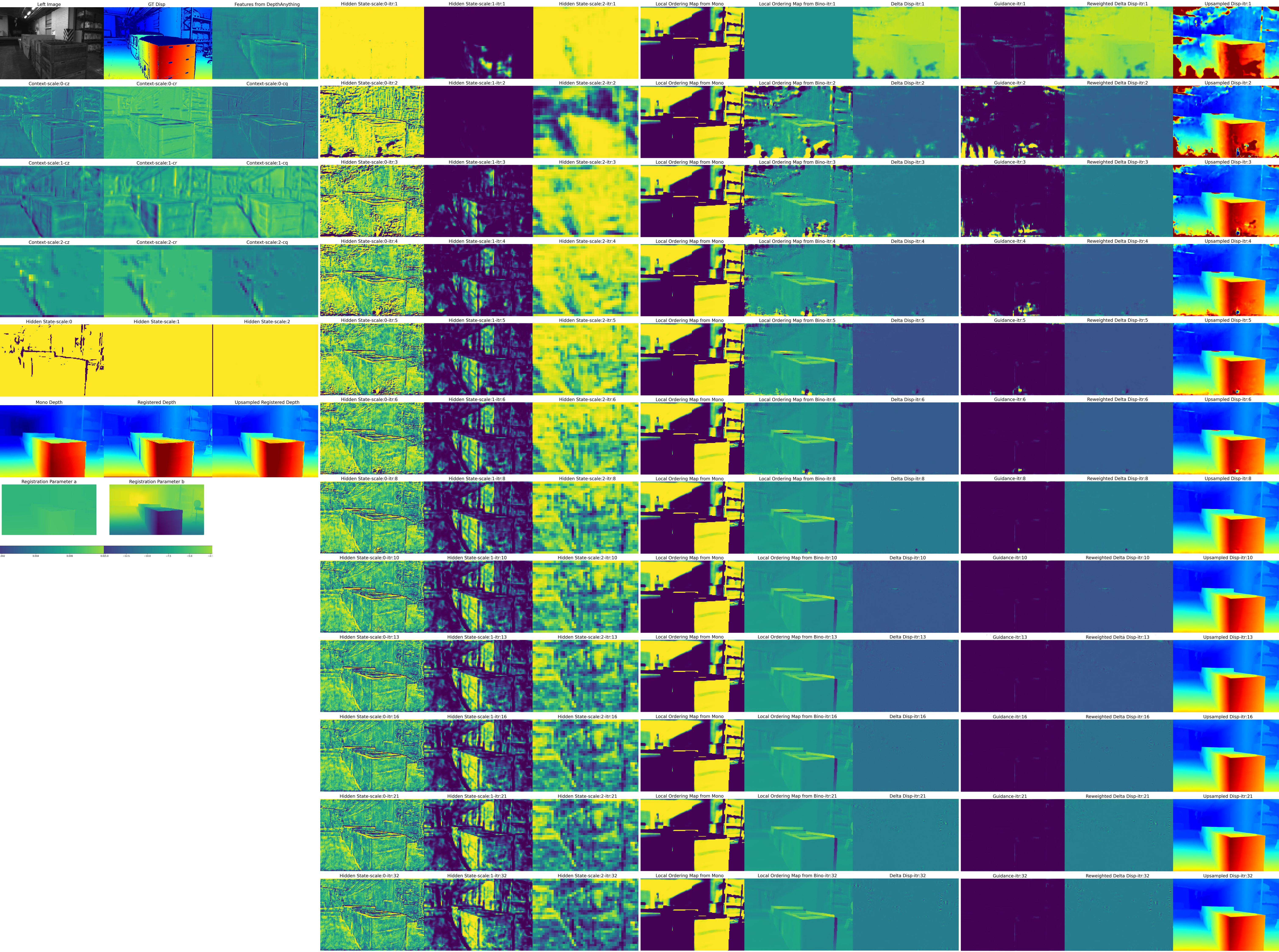}
    \caption{The visualization of intermediate results. $itr$: the current iteration. $cz, cr, cq$: context used in GRU. $scale$: scale $0\sim2$ represents resolution from high to low.}
    \label{Fig: vis-inter3}
\end{figure*}

\begin{figure*}[htbp]
\centering
    \includegraphics[width=0.9\textwidth]{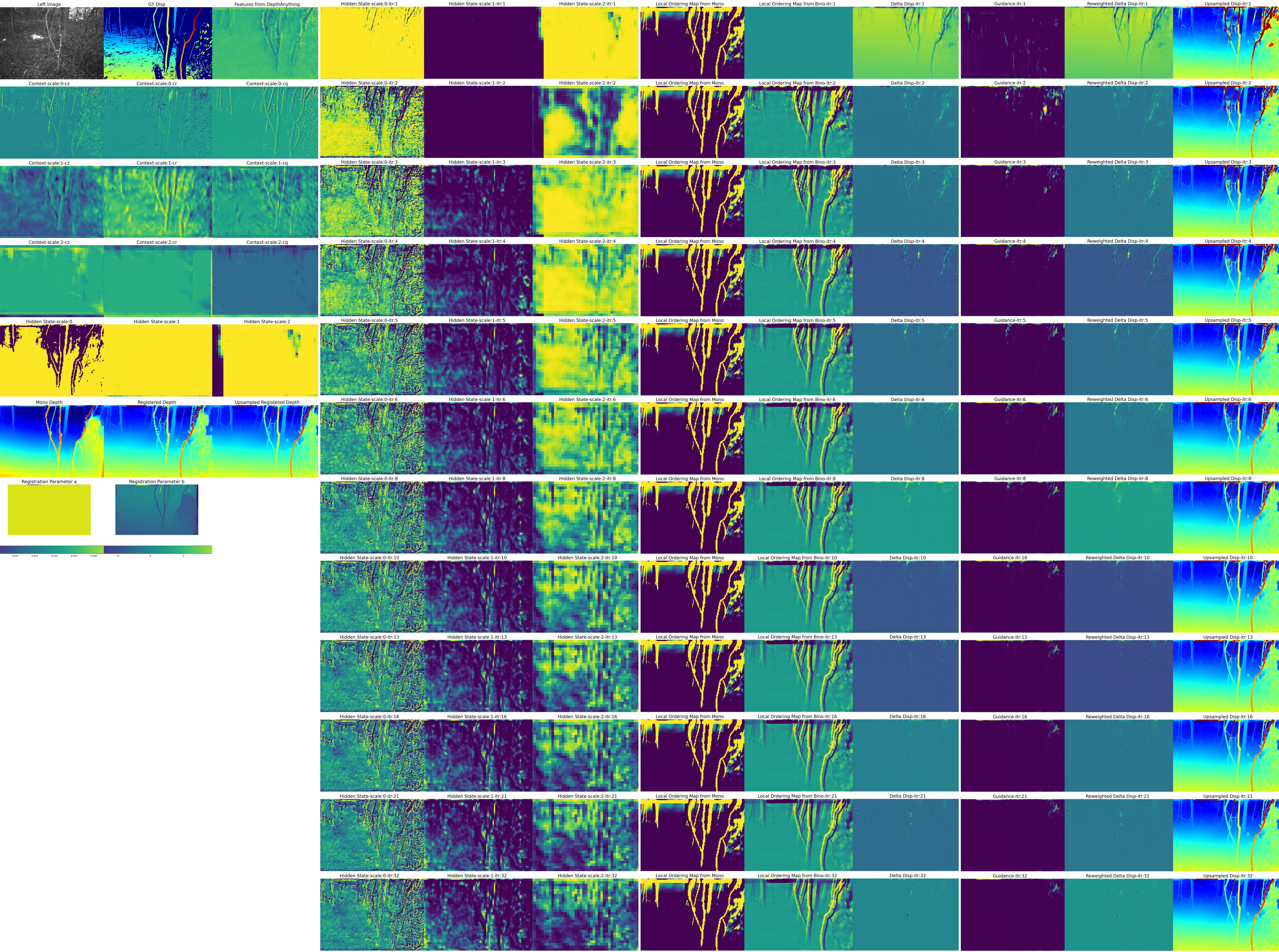}
    \caption{The visualization of intermediate results. $itr$: the current iteration. $cz, cr, cq$: context used in GRU. $scale$: scale $0\sim2$ represents resolution from high to low.}
    \label{Fig: vis-inter4}
\end{figure*}

\begin{figure*}[htbp]
\centering
    \includegraphics[width=0.9\textwidth]{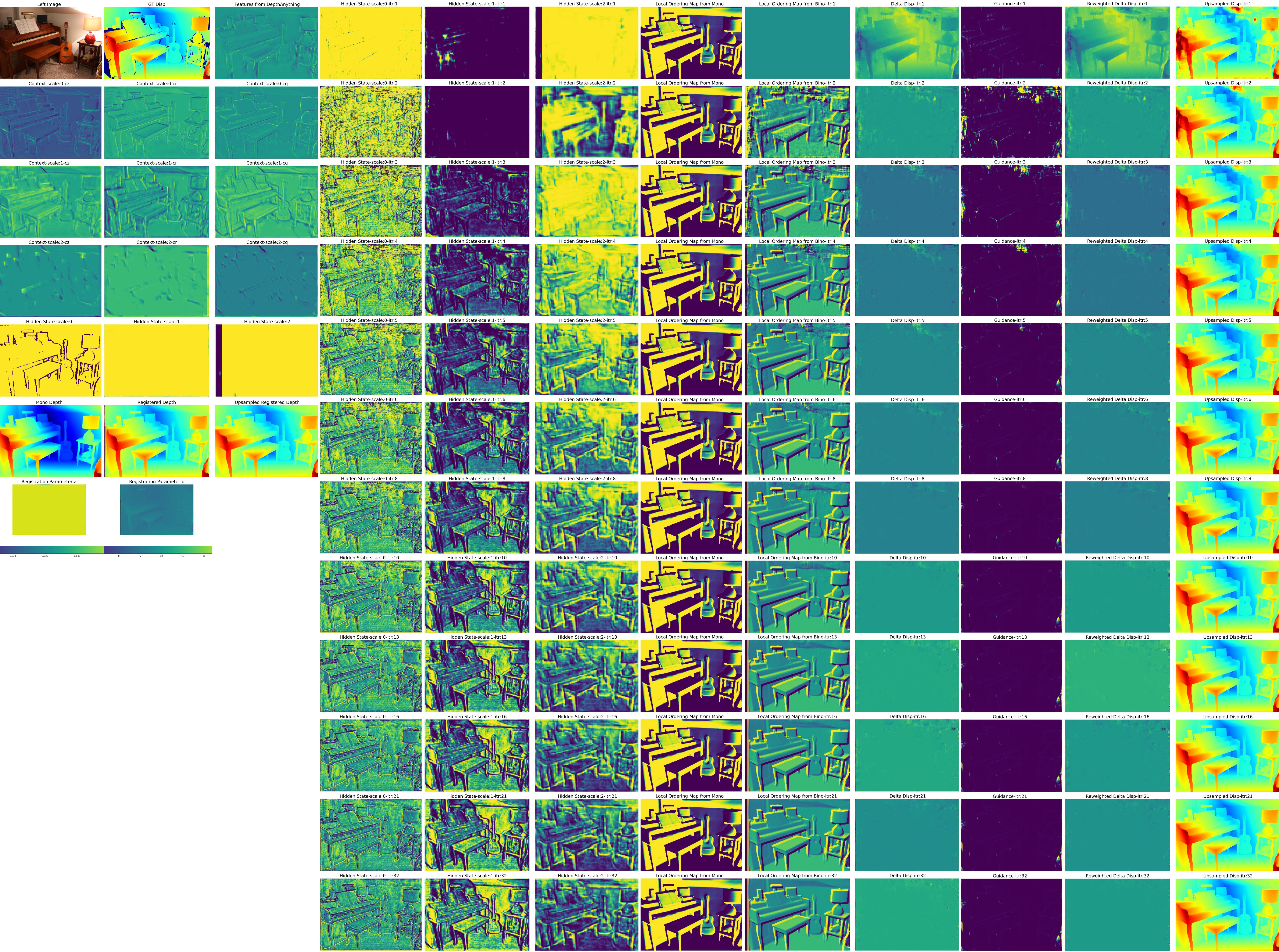}
    \caption{The visualization of intermediate results. $itr$: the current iteration. $cz, cr, cq$: context used in GRU. $scale$: scale $0\sim2$ represents resolution from high to low.}
    \label{Fig: vis-inter5}
\end{figure*}

\begin{figure*}[htbp]
\centering
    \includegraphics[width=0.9\textwidth]{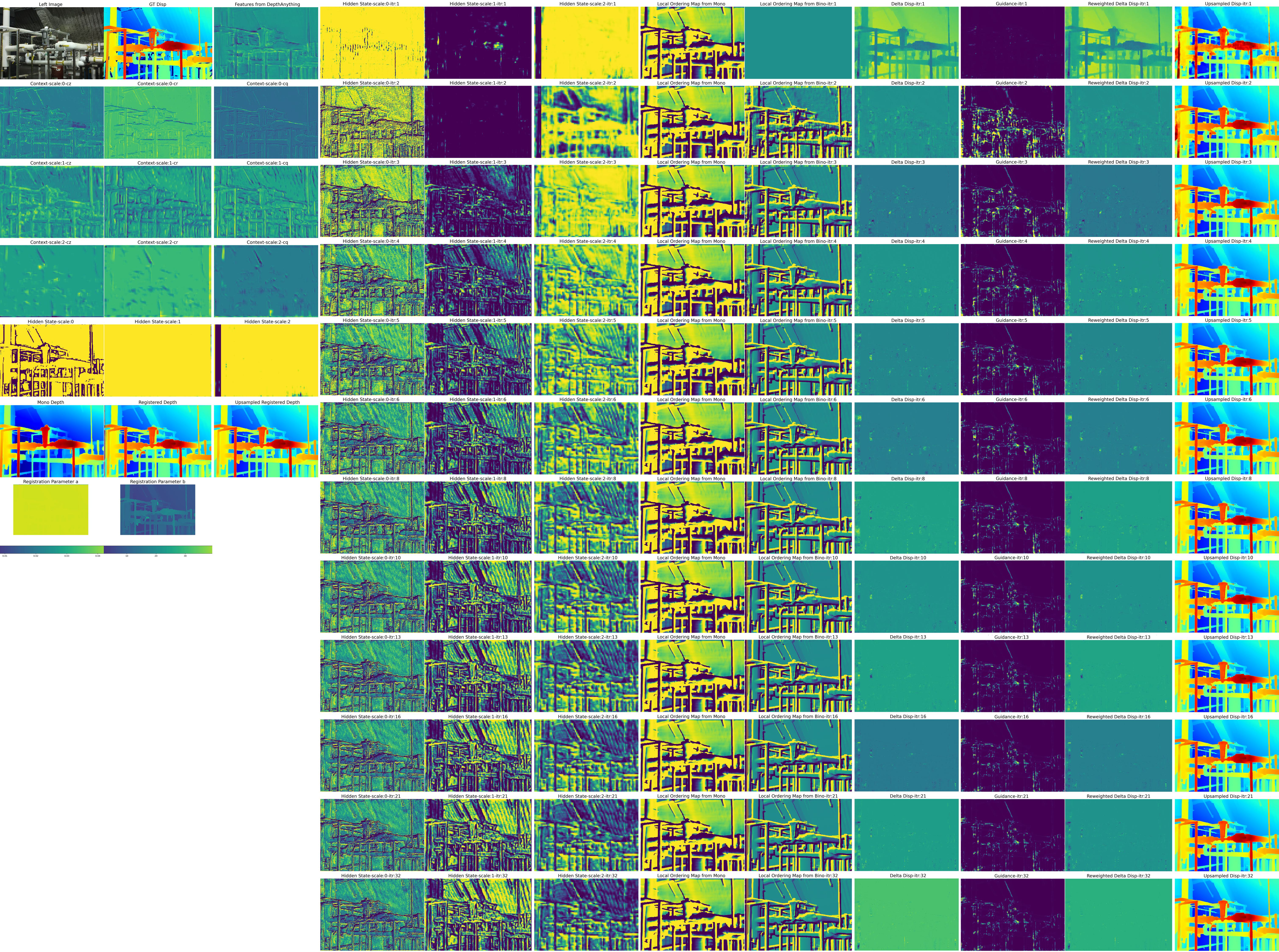}
    \caption{The visualization of intermediate results. $itr$: the current iteration. $cz, cr, cq$: context used in GRU. $scale$: scale $0\sim2$ represents resolution from high to low.}
    \label{Fig: vis-inter6}
\end{figure*}

\begin{figure*}[htbp]
\centering
    \includegraphics[width=0.9\textwidth]{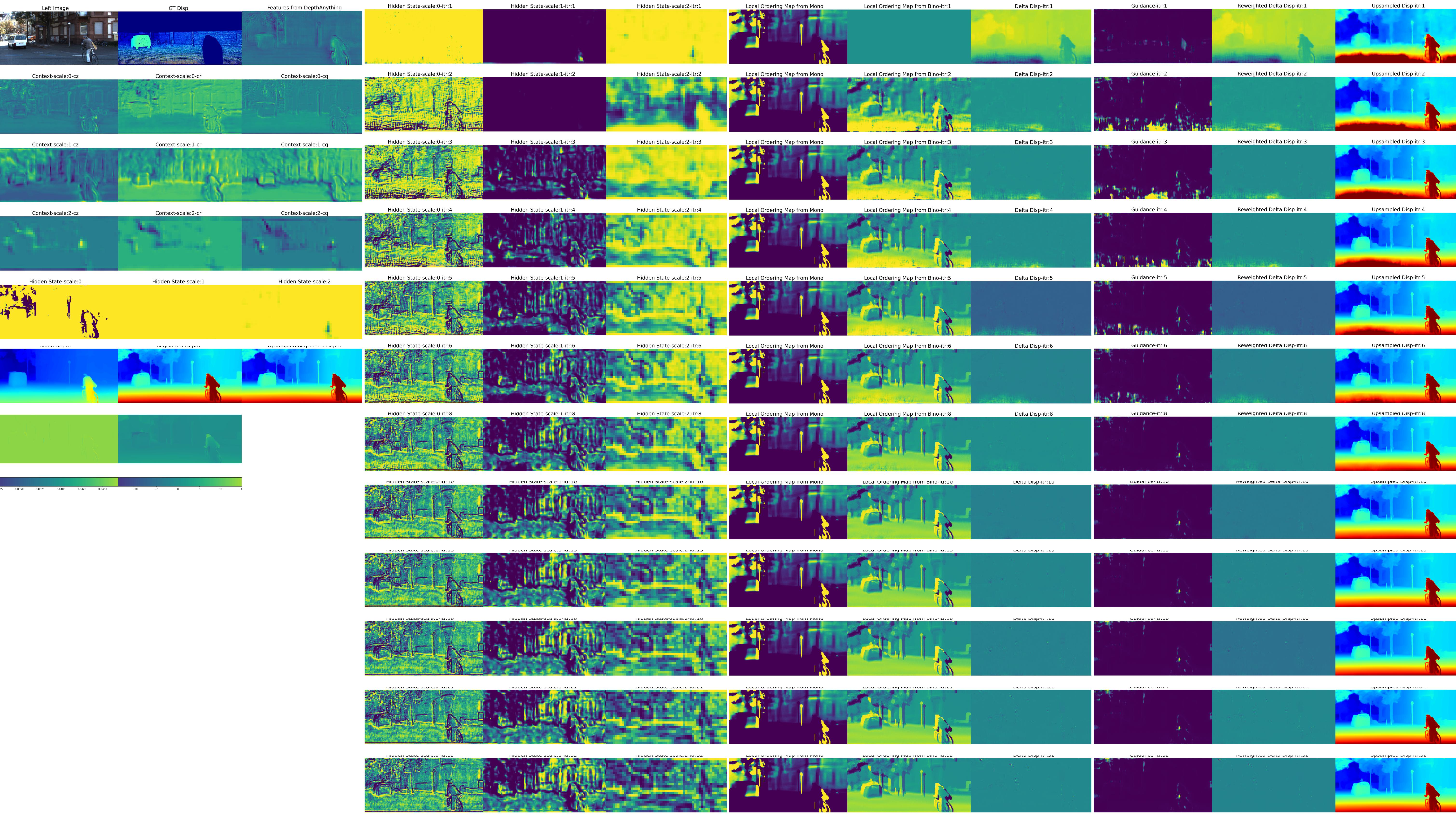}
    \caption{The visualization of intermediate results. $itr$: the current iteration. $cz, cr, cq$: context used in GRU. $scale$: scale $0\sim2$ represents resolution from high to low.}
    \label{Fig: vis-inter7}
\end{figure*}

\begin{figure*}[htbp]
\centering
    \includegraphics[width=0.9\textwidth]{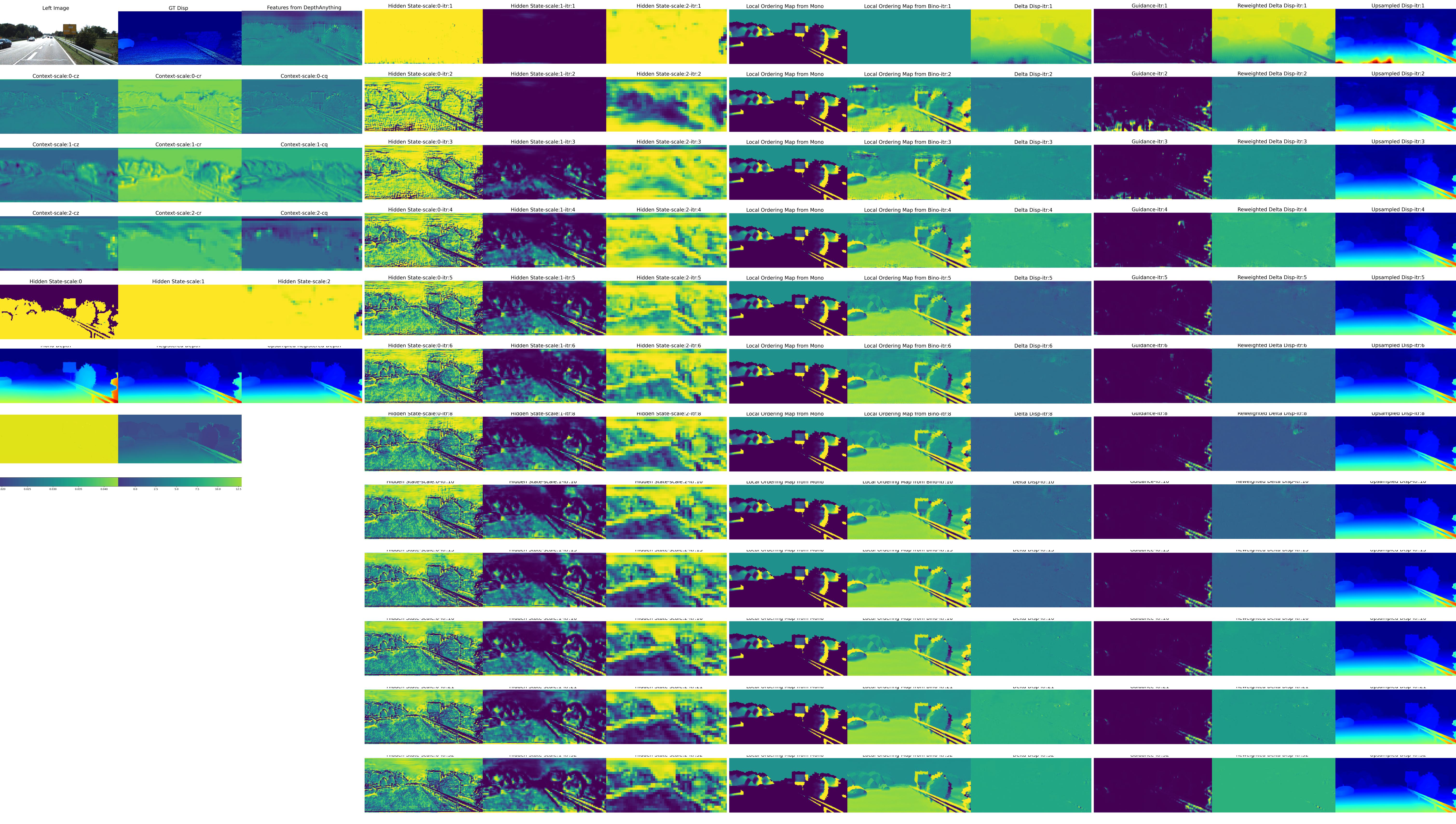}
    \caption{The visualization of intermediate results. $itr$: the current iteration. $cz, cr, cq$: context used in GRU. $scale$: scale $0\sim2$ represents resolution from high to low.}
    \label{Fig: vis-inter8}
\end{figure*}

We present more visualization for the intermediate results of global fusion in Figure \ref{Fig: vis-inter1}, \ref{Fig: vis-inter2}, \ref{Fig: vis-inter3}, \ref{Fig: vis-inter4}, \ref{Fig: vis-inter5}, \ref{Fig: vis-inter6}, \ref{Fig: vis-inter7}, and \ref{Fig: vis-inter8}. The visualization shows that the registration of monocular depth is different for each pixel, particularly on different objects. Since the monocular depth from DepthAnything is scale ambiguity but not absolute depth before registration, the visualization of it is not alinged to the ground truth range, other wise its visualization is almost a single color. The implicit learned confidence also filters out the noise of monocular depth, especially in Figure \ref{Fig: visup2}.

We provide additional visualizations of the intermediate results from global fusion in Figures \ref{Fig: vis-inter1}, \ref{Fig: vis-inter2}, \ref{Fig: vis-inter3}, \ref{Fig: vis-inter4}, \ref{Fig: vis-inter5}, \ref{Fig: vis-inter6}, \ref{Fig: vis-inter7}, and \ref{Fig: vis-inter8}. These visualizations illustrate the varying registration of monocular depth across individual pixels, particularly across different objects. Given that the monocular depth obtained from DepthAnything is scale ambiguous and does not represent absolute depth before registration, we do not align it with the ground truth range in visualization; otherwise, it would appear almost uniformly as a single color. The implicitly learned confidence also effectively filters out noise in the monocular depth as demonstrated in Figure \ref{Fig: visup2}.

\section{Future Work Discussion}
\begin{figure*}[htbp]
    \centering
    % 第一行两张图
    \begin{subfigure}[b]{0.48\textwidth}
        \centering
        \includegraphics[width=\textwidth]{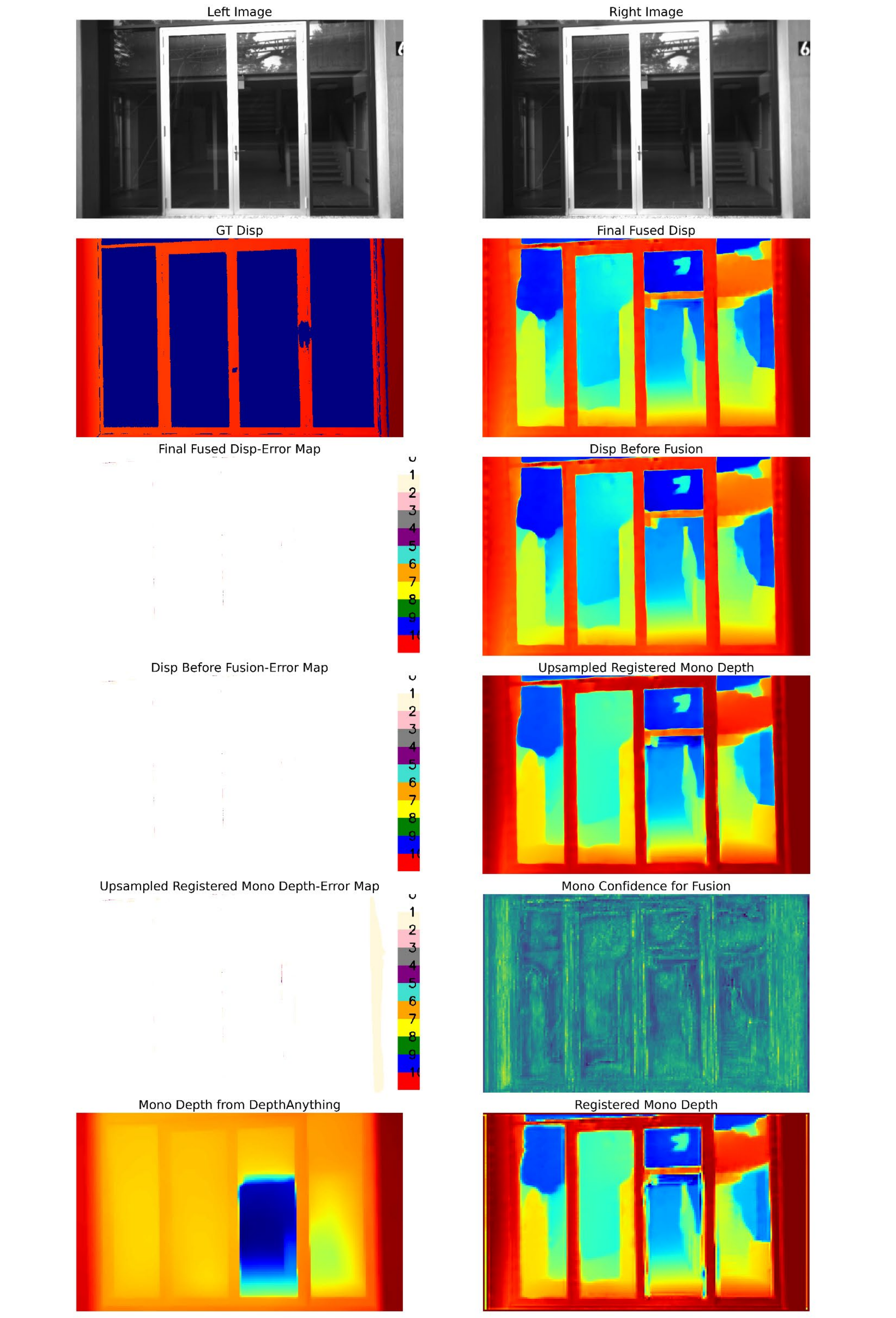} % 替换为你的图片路径
    \end{subfigure}
    % \hfill
    \begin{subfigure}[b]{0.48\textwidth}
        \centering
        \includegraphics[width=\textwidth]{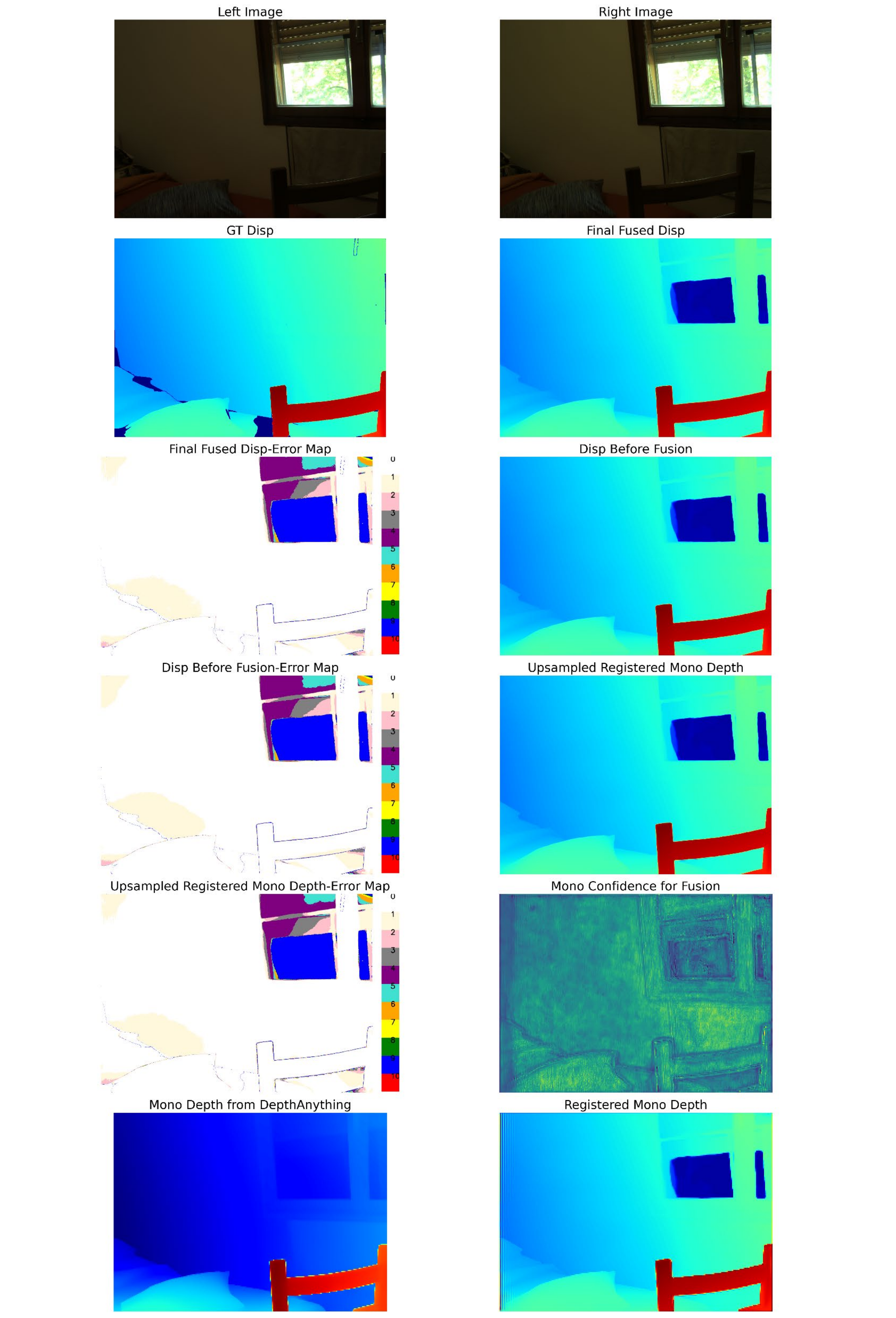} % 替换为你的图片路径
    \end{subfigure}

    \caption{The visualization for failure case analysis.}
    \label{Fig: failureCase1}
\end{figure*}

\begin{figure*}[htbp]
    \centering
    % 第二行两张图
    \begin{subfigure}[b]{0.47\textwidth}
        \centering
        \includegraphics[width=\textwidth]{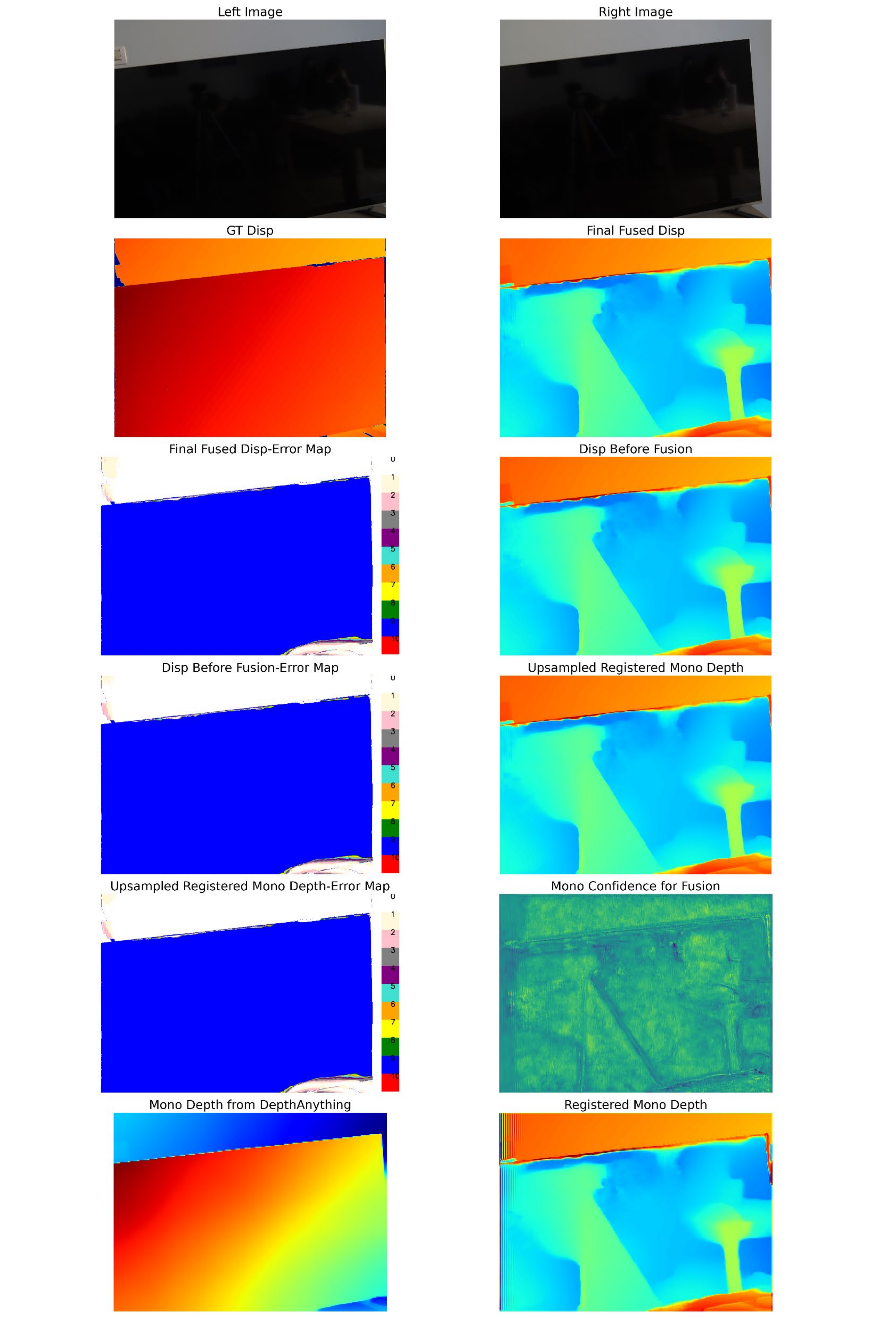} % 替换为你的图片路径
    \end{subfigure}
    % \hfill
    \begin{subfigure}[b]{0.47\textwidth}
        \centering
        \includegraphics[width=\textwidth]{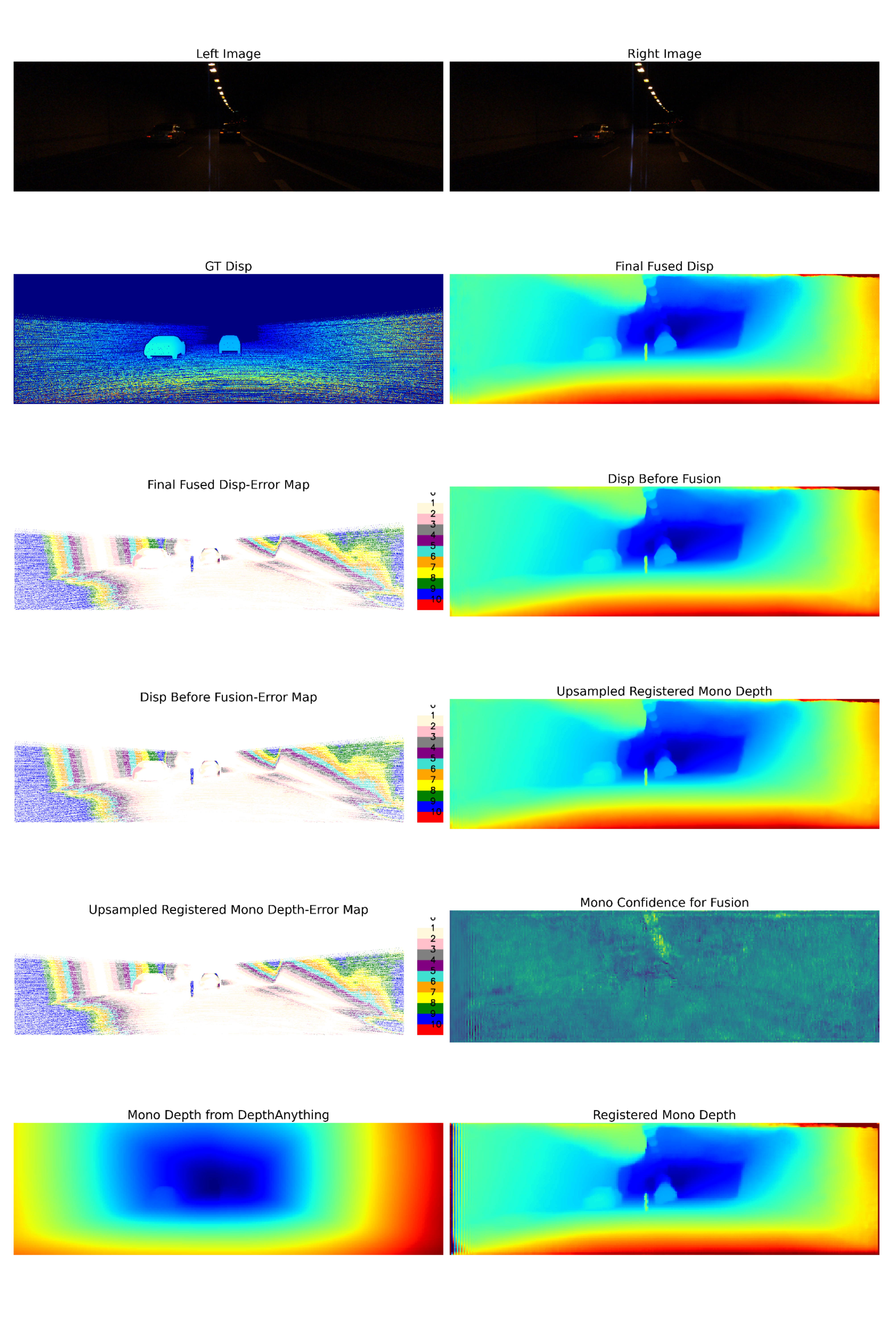} % 替换为你的图片路径
    \end{subfigure}
    \caption{The visualization for failure case analysis.}
    \label{Fig: failureCase2}
\end{figure*}

We present failure cases in Figures \ref{Fig: failureCase1} and \ref{Fig: failureCase2}. In the first failure case, our method is confused by the glass door and glass window, where both the transparent surfaces and the behind scene are significant. Unlike simple transparent objects (e.g., a glass bottle), transparent scenes raise a new challenge for robotics, as they need to perceive both the transparent surface and the scene behind it. Failure to do so may cause robots to get stuck, for instance, when trying to reach an apple behind a glass window. If the robot perceives only the glass window, it will miss the apple entirely, while perceiving only the apple means the glass acts as an unrecognized and insurmountable barrier. Therefore, a novel representation for depth estimation is necessary to allow for multiple depths at a single pixel.

In the second failure case, our method is confused by the very close black screen and the very dark tunnel. In these scenes, registering monocular depth with binocular disparity is highly challenging due to excessive and concentrated noise in the disparity, along with pixel-wise differences in monocular depth registration, particularly across different objects. Consequently, information from video streams and segmentation becomes essential, like video stereo matching or simultaneously learning segmentation.

